\documentclass[10pt,journal,compsoc]{IEEEtran}
\ifCLASSOPTIONcompsoc
  \usepackage[nocompress]{cite}
\else
  \usepackage{cite}
\fi
\ifCLASSINFOpdf
\else
\fi
\usepackage{booktabs}
\usepackage{subcaption}
\usepackage{multirow}
\usepackage{mmstyle}
\usepackage{mathtools}
\usepackage{hanging}
\usepackage{tablefootnote}
\usepackage{epsfig}
\usepackage{graphicx}
\usepackage{bmpsize}
\usepackage{amsmath}
\usepackage{amssymb}
\usepackage{url}

\usepackage{marvosym}
\usepackage{color}
\usepackage{makecell}
\usepackage{colortbl}
\definecolor{Gray}{gray}{.9}
\usepackage[breaklinks=false,colorlinks=false,bookmarks=false]{hyperref}

\begin{document}
%
\title{Reference-based Image and Video Super-Resolution via $C^{2}$-Matching}
%
%
%
%

\author{Yuming~Jiang,~Kelvin~C.K.~Chan,~Xintao~Wang,~Chen~Change~Loy,~and~Ziwei~Liu
  \IEEEcompsocitemizethanks{
    \IEEEcompsocthanksitem Yuming~Jiang is with the S-Lab, Nanyang Technological University, Singapore, 639798. E-mail: yuming002@e.ntu.edu.sg.
    \IEEEcompsocthanksitem Kelvin~C.K.~Chan is the S-Lab, Nanyang Technological University, Singapore, 639798. E-mail: chan0899@e.ntu.edu.sg.
    \IEEEcompsocthanksitem Xintao~Wang is with Applied Research Center, Tencent PCG, Shenzhen, China, 518000. E-mail: xintao.wang@outlook.com.
    \IEEEcompsocthanksitem Chen~Change~Loy is with S-Lab, Nanyang Technological University, Singapore, 639798. E-mail: ccloy@ntu.edu.sg.
    \IEEEcompsocthanksitem Ziwei~Liu is with S-Lab, Nanyang Technological University, Singapore, 639798. E-mail: ziwei.liu@ntu.edu.sg.}
}

%
%

\markboth{Journal of \LaTeX\ Class Files,~Vol.~14, No.~8, August~2015}%
{Shell \MakeLowercase{\textit{et al.}}: Bare Demo of IEEEtran.cls for Computer Society Journals}
%



\IEEEtitleabstractindextext{%

\begin{abstract}

Reference-based Super-Resolution (Ref-SR) has recently emerged as a promising paradigm to enhance a low-resolution (LR) input image or video by introducing an additional high-resolution (HR) reference image. 
Existing Ref-SR methods mostly rely on implicit correspondence matching to borrow HR textures from reference images to compensate for the information loss in input images.
However, performing local transfer is difficult because of two gaps between input and reference images: the transformation gap (\eg scale and rotation) and the resolution gap (\eg HR and LR).   
To tackle these challenges, we propose $C^{2}$-Matching in this work, which performs explicit robust matching crossing transformation and resolution.
1) To bridge the transformation gap, we propose a contrastive correspondence network, which learns transformation-robust correspondences using augmented views of the input image.
2) To address the resolution gap, we adopt teacher-student correlation distillation, which distills knowledge from the easier HR-HR matching to guide the more ambiguous LR-HR matching.   
3) Finally, we design a dynamic aggregation module to address the potential misalignment issue between input images and reference images.
In addition, to faithfully evaluate the performance of Reference-based Image Super-Resolution (Ref Image SR) under a realistic setting, we contribute the Webly-Referenced SR (WR-SR) dataset, mimicking the practical usage scenario. 
We also extend $C^{2}$-Matching to Reference-based Video Super-Resolution (Ref VSR) task, where an image taken in a similar scene serves as the HR reference image.
Extensive experiments demonstrate that our proposed $C^{2}$-Matching significantly outperforms state of the arts by up to 0.7dB on the standard CUFED5 benchmark and also boosts the performance of video super-resolution by incorporating the $C^{2}$-Matching component into Video SR pipelines.
Notably, $C^{2}$-Matching also shows great generalizability on WR-SR dataset as well as robustness across large scale and rotation transformations.
Codes and datasets are available at \url{https://github.com/yumingj/C2-Matching}.d

\end{abstract}

  }

\maketitle

\IEEEdisplaynontitleabstractindextext

%
\IEEEpeerreviewmaketitle

\section{Introduction}

\IEEEPARstart{R}{eference-based} super-resolution (Ref-SR) \cite{zheng2018crossnet, zhang2019image, yang2020learning, Shim_2020_CVPR, jiang2021robust} has attracted substantial attention in recent years. Compared to single-image super-resolution (SISR) \cite{dong2015image, kim2016accurate, kim2016deeply, lim2017enhanced, shi2016real, dai2019second} or video super-resolution (VSR) \cite{huang2015bidirectional, liu2013bayesian, takeda2009super, yi2019progressive, li2020mucan, isobe2020video, isobe2020videocvpr}, where the only input is a single low-resolution (LR) image or a sequence of video frames, Ref-SR super-resolves the LR input with the guidance of an additional high-resolution (HR) reference image. Textures of the HR reference image are transferred to provide more fine details for the LR input.

\begin{figure}
  \begin{center}
      \includegraphics[width=1.0\linewidth]{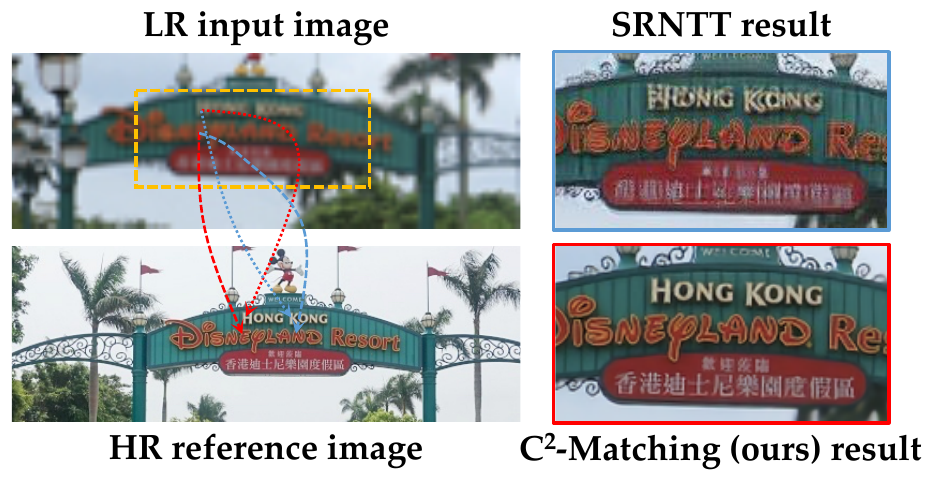}
  \end{center}
  \caption{\textbf{\textit{Cross transformation} and \textit{Cross resolution} matching are performed in our $C^{2}$-Matching}. Our proposed $C^{2}$-Matching successfully transfers the HR details of the reference image by finding more accurate correspondences. The correspondences found by our method are marked in red and the correspondences found by SRNTT \cite{zhang2019image} are marked in blue.}
  \label{teaser}
\end{figure}

The key step in texture transfer for Ref-SR is to find correspondences between the LR input and the HR reference image. Existing methods \cite{zhang2019image, yang2020learning, xiefeature} perform correspondence matching implicitly. Correspondences are firstly computed based on the content and appearance similarities, and then they are employed in the main framework.
However, it is a difficult task to accurately compute the correspondences under real-world variations due to two major challenges: \textbf{1)} the underlying transformation gap between input images and reference images; \textbf{2)} the resolution gap between input images and reference images. 
In Ref-SR, same objects or similar texture patterns are often present in both input images and reference images, but their appearances vary due to scale and rotation transformations. In this case, correspondences computed purely by appearance are inaccurate, leading to an unsatisfactory texture transfer.
For the resolution gap, due to the imbalance in the amount of information contained in an LR input and an HR reference image, the latter is often downsampled (to an LR one) to match the former (in resolution). The downsampling operation inevitably results in information loss, hampering the search for accurate correspondences, especially for the fine-texture regions \cite{liu2014fast}.

To address the aforementioned challenges, we propose $C^{2}$-matching for Robust Reference-based Super-Resolution, where \textbf{\textit{Cross transformation}} and \textbf{\textit{Cross resolution}} matching are explicitly performed. 
To handle the transformation gap, a contrastive correspondence network is proposed to learn transformation-robust correspondences between input images and reference images. Specifically, we employ a triplet margin loss to minimize the distance of point-wise features before and after transformations while maximizing the distance of irrelevant features. Thus, the extracted feature descriptors are robust to scale and rotation transformations, and can be used to compute more accurate correspondences than SRNTT \cite{zhang2019image} (red dotted lines shown in Fig.~\ref{teaser}). 

As for the resolution gap, inspired by knowledge distillation, we propose teacher-student correlation distillation. 
In particular, we train the teacher contrastive correspondence network for HR-HR matching. Since the teacher network takes two HR images as input, it is better at matching the regions with complicated textures. Thus, the knowledge of the teacher model can be distilled to guide the more ambiguous LR-HR matching.
The teacher-student correlation distillation enables the contrastive correspondence network to compute correspondences more accurately for texture regions.

After obtaining correspondences, we then fuse the information of reference images through a dynamic aggregation module to transfer the HR textures.
With $C^{2}$-Matching, we achieve 0.7dB improvement on the standard CUFED5 dataset 
\cite{zhang2019image}.
As shown in Fig.~\ref{teaser}, compared to SRNTT \cite{zhang2019image}, our $C^{2}$-Matching finds more accurate correspondences (marked as red dotted lines) and thus has a superior restoration performance.

To facilitate the evaluation of reference-based image super-resolution (Ref Image SR) task in a more realistic setting, 
we contribute a new dataset named Webly-Reference SR (WR-SR) dataset. In real-world applications, given an LR image, users may find its similar HR reference images through some web search engines. Motivated by this, for every input image in WR-SR, we search for its reference image through Google Image. The collected WR-SR can serve as a benchmark for real-world scenarios.

Apart from Ref Image SR, we also extend our proposed $C^{2}$-Matching to reference-based video super-resolution (Ref VSR) task, where an HR image taken in a similar scene to the input video is adopted as the reference image. In this work,
we demonstrate the effectiveness of our proposed $C^{2}$-Matching in the popular recurrent framework. 
Concretely, we incorporate our $C^{2}$-Matching components into BasicVSR \cite{chan2021basicvsr}. For each frame, in addition to the forward features and backward features, a reference feature fused by our dynamic aggregation module is also provided to aid the super-resolution. 
We then utilize an attention module to generate weight masks to better fuse the forward feature, backward feature and reference feature.
Compared to state-of-the-art VSR methods, with the aid of additional HR reference images, our proposed $C^{2}$-Matching further boosts the performance on REDS dataset~\cite{nah2019ntire} over 0.3dB and Vid4 dataset~\cite{liu2013bayesian} over 1dB.

To summarize, our main contributions are as follows:
\textbf{1)} To mitigate the transformation gap, we propose the contrastive correspondence network to compute correspondences more robust to scale and rotation transformations.
    %
\textbf{2)} To bridge the resolution gap, a teacher-student correlation distillation is employed to further boost the performance of student LR-HR matching model with the guidance of HR-HR matching, especially for fine texture regions.
    %
\textbf{3)} We contribute a new benchmark dataset named Webly-Referenced SR (WR-SR) to encourage a more practical application in real scenarios.
    %
\textbf{4)} To the best of our knowledge, we are the first to propose the Ref VSR task. 
We extend our proposed $C^{2}$-Matching for Ref VSR and boost the performance.

Compared with the earlier version in CVPR 2021 \cite{jiang2021robust}, we demonstrate the possibility of $C^{2}$-Matching on Ref VSR problem. Specifically, we integrate $C^{2}$-Matching into the recurrent VSR framework and utilize an attention mechanism to better fuse features. 
We evaluate the effectiveness of our proposed pipeline on VSR datasets by comparing it with VSR methods as well as two baselines adapted from Ref-SR methods. 
Apart from the methodology, the manuscript is also improved by providing more experimental analysis and technical details.

\section{Related Work}

\noindent\textbf{Single Image Super-Resolution.} Single image super-resoltion (SISR) aims to recover the HR details of LR images. The only input to the SISR task is the LR image. Dong \etal \cite{dong2015image} introduced deep learning into SISR tasks by formulating the SISR task as an image-to-image translation problem. 
Later, SR networks had gone deeper with the help of residual blocks and attention mechanisms \cite{shi2016real, lim2017enhanced, kim2016deeply, zhang2018image, zhang2018residual, ledig2017photo, dai2019second, zhou2020cross}. However, the visual quality of the output SR images did not improve. The problem was the mean square error (MSE) loss function. In order to improve the perceptual quality, perceptual loss \cite{johnson2016perceptual, sajjadi2017enhancenet}, generative loss and adversarial loss \cite{ledig2017photo,chan2021glean,chan2022glean} were introduced into the SR network \cite{wang2018esrgan, zhang2019ranksrgan}. 
Knowledge distillation mechanism was also explored to improve the SR performance in previous works \cite{gao2018image, lee2020learning}.

\noindent\textbf{Reference-based Image Super-Resolution.} Different from SISR, where no additional information is provided, the Reference-based Image Super-Resolution (Ref-SR) task \cite{yantowards, zhangtexture, zheng2018crossnet, lu2021masa} super-resolves input images by transferring HR details of reference images. 
Patch-Match method \cite{barnes2009patchmatch} was employed in earlier approaches \cite{zhang2019image, yang2020learning} to align input images and reference images. SRNTT \cite{zhang2019image} performed correspondence matching based on the off-the-shelf VGG features \cite{simonyan2014very}. Recently, learnable feature extractors are adopted \cite{yang2020learning, xiefeature}.
Even with the learnable feature extractors, the correspondences were computed purely based on contents and appearances. 
Recent work \cite{Shim_2020_CVPR} introduced Deformable Convolution Network (DCN) \cite{dai2017deformable, zhu2019deformable} to align input images and reference images. Inspired by previous work \cite{wang2019edvr, chan2021understand}, we also propose a similar module to handle the potential misalignment issue.
A concurrent work \cite{wang2021dual} extended the Ref-SR to dual-camera super-resolution, where aligned attention was proposed to bridge the transformation gap in rotation and scale.
Similarly, Zhang \etal~\cite{SelfDZSR} studied the Ref-SR in the dual-camera setting, but they dealt with real-word images via self-supervised learning.

\begin{figure*}
   \begin{center}
      \includegraphics[width=1.0\linewidth]{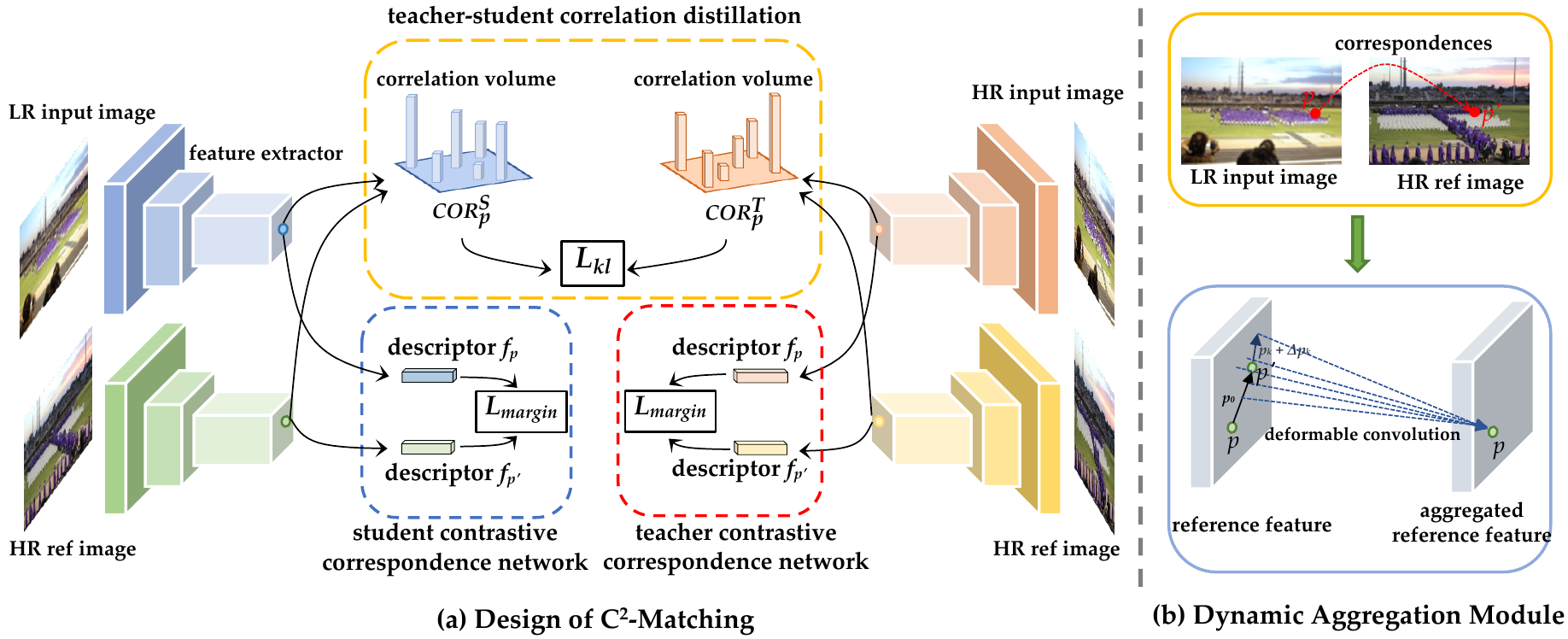}
   \end{center}
   \caption{\textbf{(a) The overview of our proposed $C^{2}$-Matching.} The contrastive correspondence network is designed for transformation-robust correspondence matching. The student contrastive correspondence network takes both the LR input image and HR reference image as input. 
   The descriptors before and after transformations are pushed closer while distances of the irrelevant descriptors are maximized by $L_{margin}$. To enable the student LR-HR contrastive correspondence network to perform correspondence matching better on highly textured regions, we embed a teacher-student correlation distillation process to distill the knowledge of the easier HR-HR teacher matching network to the student model by $L_{kl}$. \textbf{(b) The overview of dynamic aggregation module.} The correspondences computed by the trained student contrastive correspondence network are used as one component of offsets, \ie $p_{0}$. The learnable offsets $\Delta p_{k}$ are learned from input features and reference features. Information around the correspondence points in the reference feature is dynamically aggregated. }
   \label{framework_illustration}
\end{figure*}

\noindent\textbf{Video Super-Resolution.}
The objective of video super-resolution (VSR) is to upsample a given LR video sequence.
Current VSR methods \cite{huang2015bidirectional, liu2013bayesian, takeda2009super, yi2019progressive, li2020mucan, isobe2020video, isobe2020videocvpr, chan2022generalization, chan2022investigating} can be roughly divided into two categories: sliding-window based methods and recurrent methods. Sliding-window based methods aggregated neighbouring frames with a fixed window size. In earlier approaches \cite{caballero2017real, tao2017detail, xue2019video}, neighbouring frames were aligned to the intermediate frame by the predicted optical flow. Later, deformable convolutions (DCN) \cite{dai2017deformable, zhu2019deformable} were adopted to align features from neighbouring frames \cite{wang2019edvr, tian2020tdan}. 
Instead of limited neighbouring frames, recurrent frameworks \cite{isobe2020video, isobe2020revisiting, chan2021basicvsr++} leveraged information from all video frames. Isobe \etal \cite{isobe2020video} proposed to propagate the information in a forward direction via a recurrent detail structural block. In BasicVSR \cite{chan2021basicvsr}, Chan \etal proposed bi-directional propagation in a recurrent paradigm and utilized feature alignment to propagate information from other video frames.
We extend the concept of Ref-SR to VSR, where an additional reference image is provided. One LR frame could be super-resolved by fusing the information from other video frames as well as the HR reference image.
Lee \etal~\cite{lee2022reference} also studied the task of Reference-based Video Super-Resolution. In their setting, every frame has a reference frame captured from other views. In contrast, our setting only assumes one single reference image across all frames.

\noindent\textbf{Image Matching.} Scale Invariant Feature Transform (SIFT) \cite{lowe1999object} extracted local features to perform matching. With the advance of convolution neural networks (CNN), feature descriptors extracted by CNN were utilized to compute correspondences \cite{dusmanu2019d2, wiles2020d2d, duggal2019deeppruner}. Recently, SuperPoint \cite{detone2018superpoint} was proposed to perform image matching in a self-supervised manner, and graph neural network was introduced to learn feature matching \cite{sarlin2020superglue}. Needle-Match \cite{lotan2016needle} performed image matching in a more challenging setting, where two images used for matching are degraded. Different from the aforementioned methods dealing with two images of the same degradation, in our task, we focus on cross resolution matching, \ie matching between one LR image and one HR image. 
\section{Our Approach}

The overview of our proposed $C^{2}$-Matching is shown in Fig.~\ref{framework_illustration}(a). The proposed $C^{2}$-Matching consists of two major parts: 1) Contrastive Correspondence Network and 2) Teacher-Student Correlation Distillation. The contrastive correspondence network learns transformation-robust correspondence matching; the teacher-student correlation distillation transfers HR-HR matching knowledge to LR-HR matching for a more accurate correspondence matching on texture regions. To better aggregate the information of reference images, a dynamic aggregation module is proposed to handle the underlying misalignment problem  (Fig.~\ref{framework_illustration}(b)). The correspondences obtained from the contrastive correspondence network are used as the initialization of offsets in dynamic aggregation module. 
The aggregated feature can be embedded into the reference-based image super-resolution and reference-based video super-resolution.

\subsection{Contrastive Correspondence Network}

To transfer textures from reference images, correspondences should first be computed between input images and reference images, \ie specify the location of similar regions in reference images for every region in input images.

Existing methods \cite{zhang2019image, yang2020learning, xiefeature} computed correspondences according to the content and appearance similarities between degraded input images and reference images. For example, Zhang \etal \cite{zhang2019image} used VGG features \cite{simonyan2014very} for the correspondence matching while other methods \cite{yang2020learning, xiefeature} used features trained end-to-end together with the downstream tasks. 
The drawback of this scheme is that it cannot handle the matching well if there are scale and rotation transformations between input images and reference images. An inaccurate correspondence would lead to an imperfect texture transfer for restoration.
In this paper, we propose a learnable contrastive correspondence network to extract features that are robust to scale and rotation transformations. 

In the proposed contrastive correspondence network, we deal with the correspondence matching between LR input images and HR reference images. Since the resolutions of the input image and the reference image are different, we adopt two networks with the same architecture but non-shared weights for feature extractions.

For training, we synthesize HR reference images by applying homography transformations to original HR input images. By doing so, for every position $p$ in the LR input image $I$, we can compute its ground-truth correspondence point $p^\prime$ in the transformed image $I^\prime$ according to the homography transformation matrix. 
We regard point $p$ and its corresponding point $p^\prime$ as a positive pair. During optimization, we push the distances between feature representations of positive pairs closer, while maximizing the distances between other irrelevant but confusing negative pairs (defined as Eq.~\eqref{negative_pair_def}). Similar to \cite{dusmanu2019d2}, we use the triplet margin ranking loss as follows:
\begin{equation}
   L_{margin} = \frac{1}{N}\sum_{p\in {I}}\max(0, {m + \mathrm{Pos}(p) - \mathrm{Neg}(p)}),
\label{loss_margin}
\end{equation}
where $N$ is the total number of points in image $I$ and $m$ is the margin value.

The positive distance $\mathrm{Pos}(p)$ between the descriptor $f_{p}$ of position $p$ and its corresponding descriptor $f_{p^\prime}$ is defined as follows:
\begin{equation}
   \mathrm{Pos}(p) = \left \| f_{p} - f_{p^\prime} \right \|_{2}^{2}.
\end{equation}

As for the calculation of negative distance $\mathrm{Neg}(p)$, to avoid easy negative samples dominating the loss, we only select the hardest sample. The negative distance is defined as follows:
\begin{equation}
   \begin{split}
    \mathrm{Neg}(p) = \min( \min_{k \in I^\prime, \left \| k - p^\prime \right \|_\infty  > T  } \left \| f_{p} - f_{k} \right \|_{2}^{2},\\
   \min_{k \in I,  \left \| k - p \right \|_\infty > T } \left \| f_{p^\prime} - f_{k} \right \|_{2}^{2}),
   \end{split}
\label{negative_pair_def}
\end{equation}
where $T$ is a threshold to filter out neighboring points of the ground-truth correspondence point.

Thanks to the triplet margin ranking loss and the transformed versions of input image pairs, the contrastive correspondence network can generate feature descriptors more robust to scale and rotation transformations.
Once trained, the original LR input image and the HR reference image are fed into the contrastive correspondence network to compute their feature descriptors. For every position $p$ in input images, we compute its correspondence point $p^\prime$ in reference images $I^\prime$ as follows:
\begin{equation}
    p^\prime = \argmax_{q \in I^\prime} \frac{f_p}{\left \| f_p \right \|} \cdot \frac{f_q}{\left \| f_q \right \|}.
\label{corres_comp}
\end{equation}

\begin{figure*}
   \begin{center}
      \includegraphics[width=1.0\linewidth]{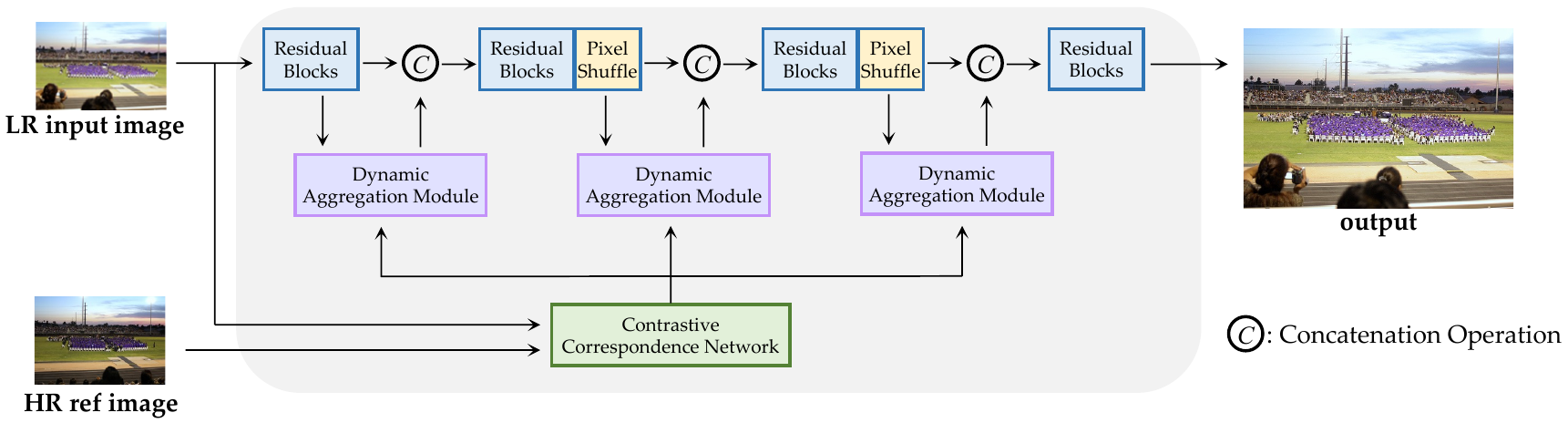}
   \end{center}
   \caption{\textbf{Pipeline of Reference-based Image Super-Resolution.} The input image is firstly fed into residual blocks to extract features. The extracted features and reference features are then fed into dynamic aggregation module to predict learnable offsets. Correspondences computed by contrastive correspondence network are used to aggregate reference features. The aggregated reference features are concatenated with features extracted by residual blocks.
   PixelShuffle is adopted to upsample the features.
   The reference features are aggregated at three scales. }
   \label{image_pipeline_illustration}
\end{figure*}

\subsection{Teacher-Student Correlation Distillation}

Since a lot of information is lost in LR input images, correspondence matching is difficult, especially for highly textured regions. 
Matching between two HR images has a better performance than LR-HR matching.
To mitigate the gap, we employ the idea of knowledge distillation \cite{hinton2015distilling} into our framework. Traditional knowledge distillation tasks \cite{hinton2015distilling, liu2019structured} deal with model compression issues, while we aim to transfer the matching ability of HR-HR matching to LR-HR matching. Thus, different from the traditional knowledge distillation models that have the same inputs but with different model capacities, in our tasks, the teacher HR-HR matching model and the student LR-HR matching have the exact same designs of architecture, but with different inputs.

The distances between the descriptors of HR input images and reference images can  provide supervision for knowledge distillation. Thus, we propose to push closer the correlation volume (a matrix that represents the distances between descriptors of input images and reference images) of teacher model to that of student model. 
For an input image, we have $N$ descriptors, and its reference image has $M$ descriptors.
By computing correlations between descriptors of input images and reference images, we can obtain an $N \times M$ matrix to represent the correlation volume, and view it as a probability distribution by applying a softmax function with temperature $\tau$ over it. To summarize, the correlation of the descriptor of input image at position $p$ and the descriptor of reference image at position $q$ is computed as follows:
\begin{equation}
  \mathrm{cor}_{pq} = \frac{e^{\frac{f_p}{\left \| f_p \right \|} \cdot \frac{f_q}{\left \| f_q \right \|} /\tau }}{\sum_{k \in I^\prime} e^{\frac{f_p}{\left \| f_p \right \|} \cdot \frac{f_k}{\left \| f_k \right \|} /\tau }}.
  \label{correlation_equa}
\end{equation}

By computing the correlations $\mathrm{cor}_{pq}$ for every pair of descriptors $p$ and $q$, 
we can obtain the correlation volume. We denote $\mathrm{COR}^{T}$ and $\mathrm{COR}^{S}$ as the teacher correlation volume and student correlation volume, respectively.
For every descriptor $p$ of input image, the divergence of teacher model's correlation and student model's correlation can be measured by Kullback Leibler divergence as follows:
\begin{equation}
\begin{split}
   \mathrm{Div}_{p} = \mathrm{KL}(\mathrm{COR}_{p}^{T} || \mathrm{COR}_{p}^{S}) \\  = \sum_{k \in I^\prime} \mathrm{cor}_{pk}^{T}\: log(\frac{\mathrm{cor}_{pk}^{T}}{\mathrm{cor}_{pk}^{S}}).
\end{split}
\end{equation}

The correlation volume contains the knowledge of relationship between descriptors.
By minimizing the divergence between two correlation volumes, the matching ability of teacher model can be transferred to the student model. This objective is defined as follows:
\begin{equation}
   L_{kl} = \frac{1}{N}\sum_{p \in I} \mathrm{Div}_{p}.
\end{equation}

With the teacher-student correlation distillation, the total loss used for training the contrastive correspondence network is:
\begin{equation}
   L = L_{margin} + \alpha _{kl} \cdot L_{kl},
\end{equation}
where $\alpha _{kl}$ is the weight for the KL-divergence loss.

\subsection{Dynamic Aggregation Module}

After obtaining correspondences using Eq.~\eqref{corres_comp}, we fuse textures from reference images by a dynamic aggregation module. 
In order to transfer the texture of reference image, we need to aggregate the information around the position $p^\prime$. We denote $p_{0}$ as the spatial difference between position $p$ and $p^\prime$, \ie $p_{0} = p^\prime - p$. Then the aggregated reference feature $y$ at position $p$ is computed by fusing original reference feature $x$ with a modified DCN as follows:
\begin{equation}
   y(p) = \sum_{k = 1}^{K} w_{k} \cdot x(p + p_{0} + \Delta p_{k}) \cdot \Delta m_{k},
\label{DCN_eq}
\end{equation}
where $w_{k}$ denotes the convolution kernel weight, $\Delta p_{k}$ and $\Delta m_{k}$ denote the learnable offset and modulation scalar, respectively.

The learnable offset $\Delta p_{k}$ and the modulation scalar $\Delta m_{k}$ are predicted by concatenating input features and reference features.

Compared to the reference feature aggregation operation used in \cite{zhang2019image} that cropped patches with a fixed size around corresponding points, our dynamic aggregation module dynamically utilizes the information around the precomputed corresponding points with learnable offsets $\Delta p_{k}$.

\subsection{Reference-based Image Super-Resolution}

The pipeline for reference-based image super-resolution (Ref Image SR) is shown in Fig.~\ref{image_pipeline_illustration}. The input LR image is firstly fed into residual blocks consisting of several convolution layers. The obtained feature is further enhanced by transferring aggregated reference features at different scales. At scale $l$, the aggregated reference feature $R_{l-1}$ is transferred as follows: 
\begin{equation}
F_l = \text{Res} \left( F_{l-1} \: \| \: R_{l-1} \right)  + F_{l-1}
\end{equation}
where $\text{Res}(\cdot)$ denotes residual blocks, $\|$ denotes concatenations.

The texture transfer operation is followed by a PixelShuffle operation \cite{shi2016real} to upscale the features by $2 \times$.
The reference features are aggregated at three levels with different resolutions (\ie pretrained VGG $\text{relu3\_1}$, $\text{relu2\_1}$, $\text{relu1\_1}$ features). The offsets used for aggregation consist of correspondences computed by contrastive correspondence network and learned offsets.

We employ the commonly used reconstruction loss $L_{rec}$, perceptual loss $L_{per}$, and adversarial loss $L_{adv}$ for the image super-resolution network. 

\noindent\textbf{Reconstruction Loss.} The $\ell_1$-norm is adopted to keep the spatial structure of the LR images. It is defined as follows:
\begin{equation}
   L_{rec} = \left \| I^{HR} - I^{SR} \right \|_{1}.
\end{equation}

\noindent\textbf{Perceptual Loss.} The perceptual loss \cite{johnson2016perceptual} is employed to improve the visual quality. It is defined as follows:
\begin{equation}
    L_{per} = \frac{1}{V}\sum_{i=1}^{C}\left \| \phi _{i} (I^{HR}) - \phi _{i} (I^{SR})  \right \|_{F},
\end{equation}
where $V$ and $C$ denote the volume and channel number of feature maps. $\phi$ denotes the relu5\_1 features of VGG19 model \cite{simonyan2014very}. $\left \| \cdot  \right \|_{F}$ denotes the Frobenius norm.

\noindent\textbf{Adversarial Loss.} The adversarial loss \cite{ledig2017photo} is defined as follows:
\begin{equation}
    L_{adv} = - D(I^{SR}).
\end{equation}
The loss for training discriminator $D$ is defined as follows:
\begin{equation}
    L_{D} = D(I^{SR}) - D(I^{HR}) + \lambda (\left \| \nabla _{\hat{I}} D(\hat{I}) \right \|_{2} - 1)^2.
\end{equation}
where $\hat{I}$ is the random convex combination of $I^{SR}$ and $I^{HR}$.

\begin{figure*}
  \begin{center}
      \includegraphics[width=1.0\linewidth]{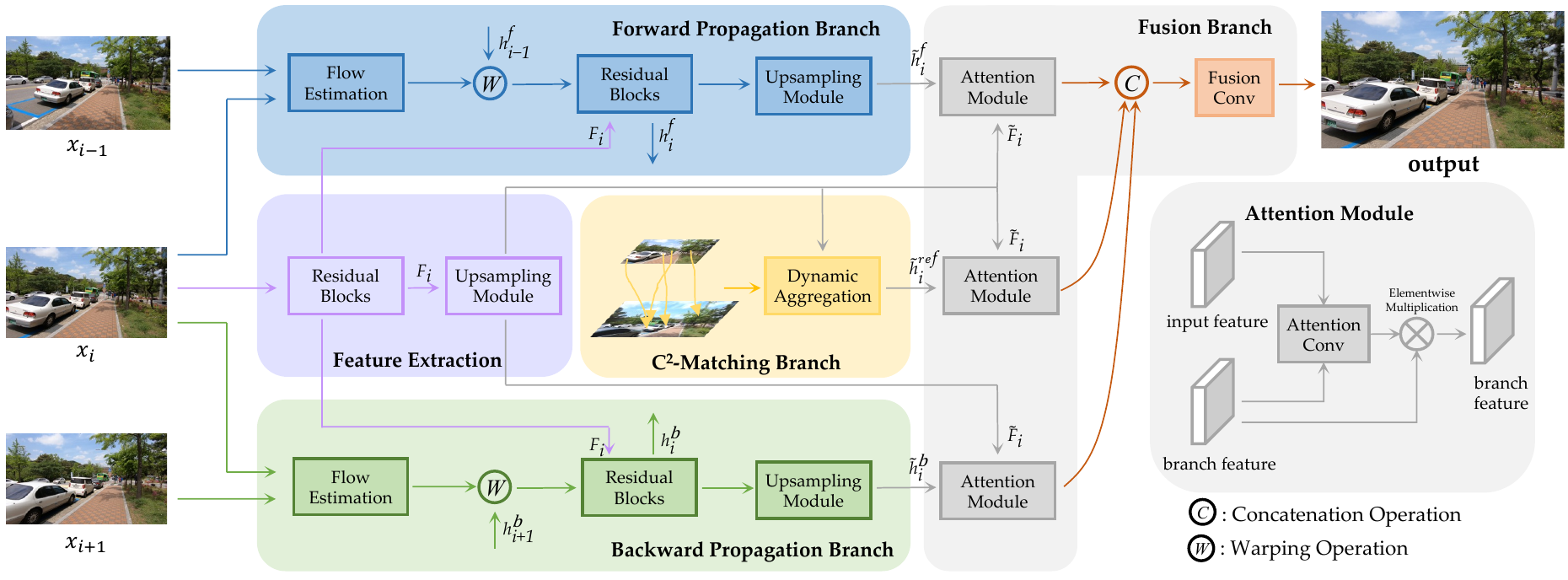}
  \end{center}
  \vspace{5pt}
  \caption{\textbf{Pipeline of Reference-based Video Super-Resolution.} The restoration of one frame in video consists of five components: feature extraction, forward propagation branch, $C^{2}$-Matching branch, backward propagation branch, and fusion branch. In the forward propagation branch, the feature propagated from the previous frames is warped by the estimated flow. The backward propagation branch is designed in a similar paradigm. In $C^{2}$-Matching branch, correspondences are firstly computed between the current frame and HR reference image. Then the reference features are aggregated according to the correspondences and learnable offsets. For the feature generated by each branch, an attention module is employed to generate a mask that assigns higher weights to more similar regions and lower weights to dissimilar regions. Finally, the weighted features maps from three branches are concatenated together and then fed into the fusion branch to generate the final output.}
  \label{video_pipeline_illustration}
  \vspace{5pt}
\end{figure*}

\subsection{Reference-based Video Super-Resolution}

The pipeline for reference-based video super-resolution (Ref VSR) is shown in Fig.~\ref{video_pipeline_illustration}. Inspired by BasicVSR~\cite{chan2021basicvsr}, we use the bi-directional recurrent framework to restore an LR video. The whole pipeline consists of five components: feature extraction, forward propagation branch, backward propagation branch, $C^{2}$-Matching branch, and fusion branch.

\noindent\textbf{Feature Extraction.} For each frame, residual blocks are employed to extract the feature $F_{i}$. The extracted feature $F_{i}$ will be used in the following propagation components. The feature is also upsampled $4 \times$ by the Upsamping Module, which consists of two PixelShuffle operations. The upsampled input feature $\tilde{F}_{i}$ will be used in the $C^{2}$-Matching branch and the Fusion part.  

\noindent\textbf{Forward Propagation Branch.} The forward feature is recurrently warped from the first frame using the estimated flow between neighbouring frames. After the warping operation, the feature is fed into the residual blocks:
\begin{equation}
    h_{i}^{f} = \text{Res}[F_{i}\: \| \: W (h_{i-1}^{f}, s_{i}^{f})],
\end{equation}
where $\text{W}$ denote the spatial warping operation, and $s_{i}^{f}$ is the flow estimated from the current frame $x_{i}$ and the previous frame $x_{i-1}$.

An upsampling module is then applied to the forward feature $h_{i}^{f}$. The upsampled feature $\tilde{h}_{i}^{f}$ will be used in Fusion module.

\noindent\textbf{Backward Propagation Branch.} The backward propagation branch adopts a similar design to the forward propagation branch. The backward feature $h_{i}^{b}$ is propagated starting from the last frame.

\noindent\textbf{$C^{2}$-Matching Branch.} For every frame, correspondences are computed against the HR reference image. The aggregated reference feature $\tilde{h}_{i}^{ref}$ is obtained from dynamic aggregation module, where the computed correspondences and learned offsets are used to aggregate the reference feature. In Ref-VSR task, we only aggregate the reference feature at the largest scale (\ie only $\text{relu1\_1}$ VGG features are used). 

\noindent\textbf{Fusion Branch.} To better fuse the features from different branches, an attention module is proposed. The fusion takes the similarity into consideration. For similar regions, the fusion weight should be higher. The attention module takes the upsampled input feature $\tilde{F}_{i}$ and $\tilde{h}_{i}$ as inputs and generates an attention map with the same size as the feature map $\tilde{h}_{i}$. The attention map is elementwise multiplied to the feature map.
\begin{equation}
    \tilde{h}_{i} = f (\tilde{F}_{i} \: \| \: \tilde{h}_{i}) \otimes \tilde{h}_{i},
\end{equation}
where $f(\cdot)$ denotes the attention map generation operation, which consists of several convolution layers.

\noindent\textbf{Loss Function.} For video super-resolution, we use Charbonnier loss as follows:

\begin{equation}
    L = \frac{1}{N}\sum_{i=0}^{N} \rho (y_{i} - z_{i}),
\end{equation}
where $\rho (x) = \sqrt{ x^2 + \epsilon ^2 }$, $\epsilon = 1 \times 10^{-8}$, $z_{i}$ denotes the ground-truth frame, and $N$ is the number of pixels.

\begin{figure*}
   \begin{center}
      \includegraphics[width=1.0\linewidth]{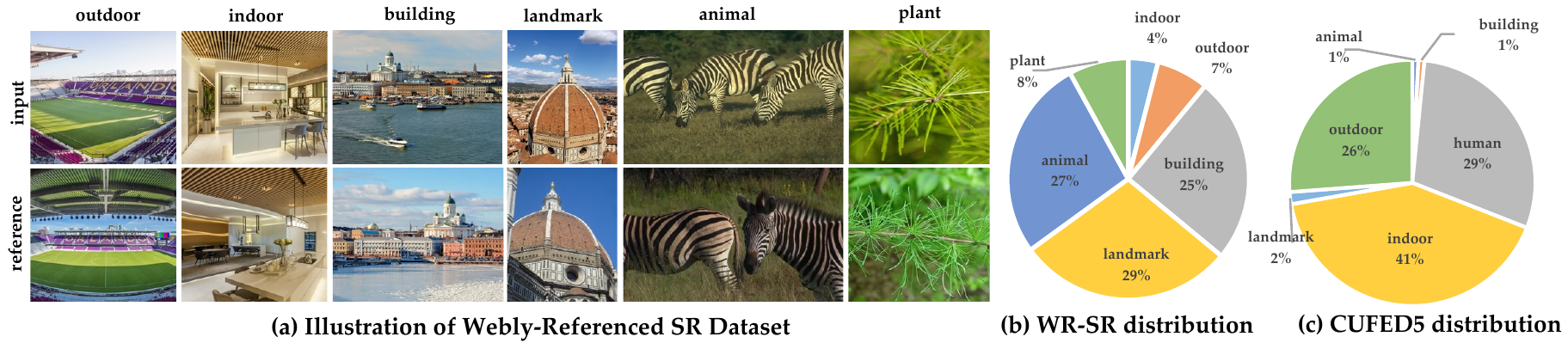}
   \end{center}
   \caption{\textbf{(a) Illustration of Webly-Referenced SR dataset.} The contents of WR-SR dataset include outdoor scenes, indoor scenes, buildings, famous landmarks, animals, and plants. The first line shows the HR input images and the second line is their reference images searched by Google Image. \textbf{(b) WR-SR Dataset Distribution. (c) CUFED5 Dataset Distribution.} The WR-SR dataset contains a more diverse category of images. It has more animal, building, and landmark images than CUFED5 dataset. }
   \vspace{5pt}
   \label{dataset_illustration}
\end{figure*}

\subsection{Implementation Details}

The overall network is trained in two-stage: \textbf{1)} training of $C^{2}$-Matching, \ie contrastive correspondence network accompanied with teacher-student correlation distillation. \textbf{2)} training of restoration network.

\noindent\textbf{Training of $C^{2}$-Matching.} We synthesize the image pairs by applying synthetic homography transformation to input images.
Homography transformation matrix is obtained by \textit{cv2.getPerspectiveTransform}.
To enable teacher-student correlation distillation, a teacher contrastive correspondence network should be first trained. The hyperparameters for the training of teacher model are set as follows: the margin value $m$ is $1.0$, the threshold value $T$ is $4.0$, the batch size is set as $8$, and the learning rate is $10^{-3}$. We use the pretrained weights of VGG-16 to initialize the feature extractor. Then the student contrastive correspondence network is trained with the teacher network fixed. The margin value $m$, threshold value $T$, batch size, and learning rate are the same as the teacher network. The temperature value $\tau$ is $0.15$, and the weight $\alpha_{kl}$ for KL-divergence loss is $15$. 

\noindent\textbf{Training of Ref Image SR.} In this stage, correspondences obtained from the student contrastive correspondence network are used for the calculation of $p_{0}$ specified in Eq.~\eqref{DCN_eq}. When computing correspondences, we use the feature descriptors of a $3 \times 3$ patch. The weights for $L_{rec}$, $L_{per}$ and $L_{adv}$ are $1.0$, $10^{-4}$ and $10^{-6}$, respectively. The learning rate for the training of restoration network is set as $10^{-4}$. For the training of the network with adversarial loss and perceptual loss, we adopt the same setting as \cite{zhang2019image} (\ie the network is trained with only reconstruction loss for the first 10K iterations).
During training, the input sizes for LR images and HR reference images are $40 \times 40$ and $160 \times 160$, respectively.
We set $K = 9$ in Eq.~(\ref{DCN_eq}).

\noindent\textbf{Training of Ref VSR.} The learning rate for flow estimation is set as $2.5 \times 10^{-5}$, and the learning rate for the rest part is set as $2 \times 10^{-4}$. The weight of flow estimation module is initialized using pretrained SPyNet~\cite{ranjan2017optical} and is fixed during the first $5,000$ iterations. The batch size is 8 and the path size is $64 \times 64$.
We set $K = 9$ in Eq.~(\ref{DCN_eq}).
\section{Webly-Referenced SR Dataset}
\label{section_dataset}

In Ref-SR tasks, the performance relies on similarities between input images and reference images. Thus, the quality of reference images is vital. Currently, Ref-SR methods are trained and evaluated on the CUFED5 dataset \cite{zhang2019image}, where each input image is accompanied by five references of different levels of similarity to the input image. 
A pair of input and reference images in CUFED5 dataset is selected from a same event album. Constructing image pairs from albums ensures a high similarity between the input and reference image. However, in realistic settings, it is not always the case that we can find the reference images from off-the-shelf albums.

In real-world applications, given an LR image, users may find possible reference images through web search engines like Google Image. Motivated by this, we propose a more reasonable dataset named Webly Referenced SR (WR-SR) Dataset to evaluate Ref-SR methods. The WR-SR dataset is much closer to the practical usage scenarios, and it is set up as follows: 

\noindent\textbf{Data Collection.} We select about 150 images from BSD500 dataset \cite{MartinFTM01} and Flickr website. These images are used as query images to search for their visually similar images through Google Image. For each query image, the top 50 similar images are saved as reference image pools for the subsequent Data Cleaning procedure.

\noindent\textbf{Data Cleaning.} Images downloaded from Google Image are of different levels of quality and similarity. Therefore, we manually select the most suitable reference image for each query image. Besides,  since some reference images are significantly larger than input images, we rescale the reference images to a comparable scale as HR input images. We also abandon the images with no proper reference images found.

\noindent\textbf{Data Organization.} A total of 80 image pairs are collected for WR-SR dataset. Fig. \ref{dataset_illustration} (a) illustrates our WR-SR dataset. The contents of the input images in our dataset include outdoor scenes, indoor scenes, building images, famous landmarks, animals and plants. We analyze the distributions of our WR-SR dataset and CUFED5 dataset. As shown in Fig.~\ref{dataset_illustration} (b), compared to CUFED5 dataset (Fig.~\ref{dataset_illustration} (c)), we have a more diverse category, and we include more animal, landmark, building and plant images.

To summarize, our WR-SR dataset has two advantages over CUFED5 dataset: \textbf{1)} The pairs of input images and reference images are collected in a more realistic way. \textbf{2)} Our contents are more diverse than CUFED5 dataset.

\begin{table*}[t]
\centering
	\caption{\small{Quantitative Comparisons on Ref Image SR (PSNR / SSIM). We group methods by SISR and Ref-SR. We mark the best results \textbf{in bold}. The models trained with GAN loss are marked in gray. The suffix `-$rec$' means only reconstruction loss is used for training.}}
\vspace{-6pt}
\addtolength{\tabcolsep}{6pt}
\small{
\begin{tabular}{c|l|ccccc}
\Xhline{1pt}
        & Method        & CUFED5        & Sun80   & Urban100  & Manga109     & WR-SR    \\ \hline
\multirow{7}{*}{SISR}   & SRCNN \cite{dong2015image}    & 25.33 / .745    & 28.26 / .781    & 24.41 / .738  & 27.12 / .850   & 27.27 / .767   \\
  & EDSR \cite{lim2017enhanced}   & 25.93 / .777     & 28.52 / .792   & 25.51 / .783   & 28.93 / .891    & 28.07 / .793   \\
  & RCAN \cite{zhang2018image}    & 26.06 / .769     & 29.86 / .810   & 25.42 / .768   & 29.38 / .895    & 28.25 / .799   \\
  & \cellcolor{Gray}SRGAN \cite{ledig2017photo}  & \cellcolor{Gray}24.40 / .702   & \cellcolor{Gray}26.76 / .725    & \cellcolor{Gray}24.07 / .729   & \cellcolor{Gray}25.12 / .802     & \cellcolor{Gray}26.21 / .728   \\
  & ENet \cite{sajjadi2017enhancenet}   & 24.24 / .695  & 26.24 / .702     & 23.63 / .711    & 25.25 / .802   & 25.47 / .699   \\
  & \cellcolor{Gray}ESRGAN \cite{wang2018esrgan}   & \cellcolor{Gray}21.90 / .633    & \cellcolor{Gray}24.18 / .651   & \cellcolor{Gray}20.91 / .620   & \cellcolor{Gray}23.53 / .797    & \cellcolor{Gray}26.07 / .726    \\
  & \cellcolor{Gray}RankSRGAN \cite{zhang2019ranksrgan}    & \cellcolor{Gray}22.31 / .635    & \cellcolor{Gray}25.60 / .667   &  \cellcolor{Gray}21.47 / .624    & \cellcolor{Gray}25.04 / .803     & \cellcolor{Gray}26.15 / .719      \\ \hline \hline
\multirow{11}{*}{Ref-SR} & CrossNet \cite{zheng2018crossnet}    & 25.48 / .764     & 28.52 / .793     & 25.11 / .764    & 23.36 / .741    & 23.77 / .612    \\ 
  & \cellcolor{Gray}SRNTT   & \cellcolor{Gray}25.61 / .764    & \cellcolor{Gray}27.59 / .756    & \cellcolor{Gray}25.09 / .774     & \cellcolor{Gray}27.54 / .862    & \cellcolor{Gray}26.53 / .745       \\
   & SRNTT-$rec$ \cite{zhang2019image}   & 26.24 / .784   & 28.54 / .793   & 25.50 / .783    & 28.95 / .885   & 27.59 / .780     \\
   & \cellcolor{Gray}TTSR   & \cellcolor{Gray}25.53 / .765     & \cellcolor{Gray}28.59 / .774     & \cellcolor{Gray}24.62 / .747    & \cellcolor{Gray}28.70 / .886     & \cellcolor{Gray}26.83 / .762  \\
   & TTSR-$rec$ \cite{yang2020learning}  & 27.09 / .804    & 30.02 / .814     & 25.87 / .784   & 30.09 / .907    & 27.97 / .792   \\
   & \cellcolor{Gray}SSEN    & \cellcolor{Gray}25.35 / .742   & \cellcolor{Gray}-  & \cellcolor{Gray}-   & \cellcolor{Gray}-  & \cellcolor{Gray}-   \\
   & SSEN-$rec$ \cite{Shim_2020_CVPR}  & 26.78 / .791    & -  & -   & -   & -   \\
   & \cellcolor{Gray}E2ENT$^{2}$    & \cellcolor{Gray}24.01 / .705   & \cellcolor{Gray}28.13 / .765   & \cellcolor{Gray}-  & \cellcolor{Gray}-  & \cellcolor{Gray}-  \\
   & E2ENT$^{2}$-$rec$ \cite{xiefeature} & 24.24 / .724   & 28.50 / .789      & -   & -   & -       \\
   & \cellcolor{Gray}CIMR   & \cellcolor{Gray}26.16 / .781    & \cellcolor{Gray}29.67 / .806   & \cellcolor{Gray}25.24 / .778   & \cellcolor{Gray}-    & \cellcolor{Gray}-   \\
   & CIMR-$rec$ \cite{yantowards}   & 26.35 / .789   & 30.07 / .813   & 25.77 / \textbf{.792} & -    & - \\  
  & \cellcolor{Gray}MASA-SR & \cellcolor{Gray}24.92 / .729 & \cellcolor{Gray}27.12 / .708 & \cellcolor{Gray}23.78 / .712 & \cellcolor{Gray}27.23 / .844 & \cellcolor{Gray}25.49 / .710 \\ 
  & MASA-SR - $rec$ \cite{lu2021masa} & 27.54 / .814 & 30.15 / .815 & \textbf{26.09} / .786 & 30.19 / .908 & 28.09 / .794 \\ \hline \hline 
\multirow{2}{*}{Ours}    & \cellcolor{Gray}$C^{2}$-Matching    & \cellcolor{Gray}27.16 / .805   & \cellcolor{Gray}29.75 / .799    & \cellcolor{Gray}25.52 / .764   & \cellcolor{Gray}29.73 / .893   & \cellcolor{Gray}27.80 / .780   \\
   & $C^{2}$-Matching-$rec$   & \textbf{28.24} / \textbf{.841} & \textbf{30.18} / \textbf{.817} & 26.03 / .785 & \textbf{30.47} / \textbf{.911} & \textbf{28.32} / \textbf{.801} \\
\Xhline{1pt}
\end{tabular}}
\label{quan_comp}
\end{table*}

\begin{figure*}
  \begin{center}
      \includegraphics[width=1.0\linewidth]{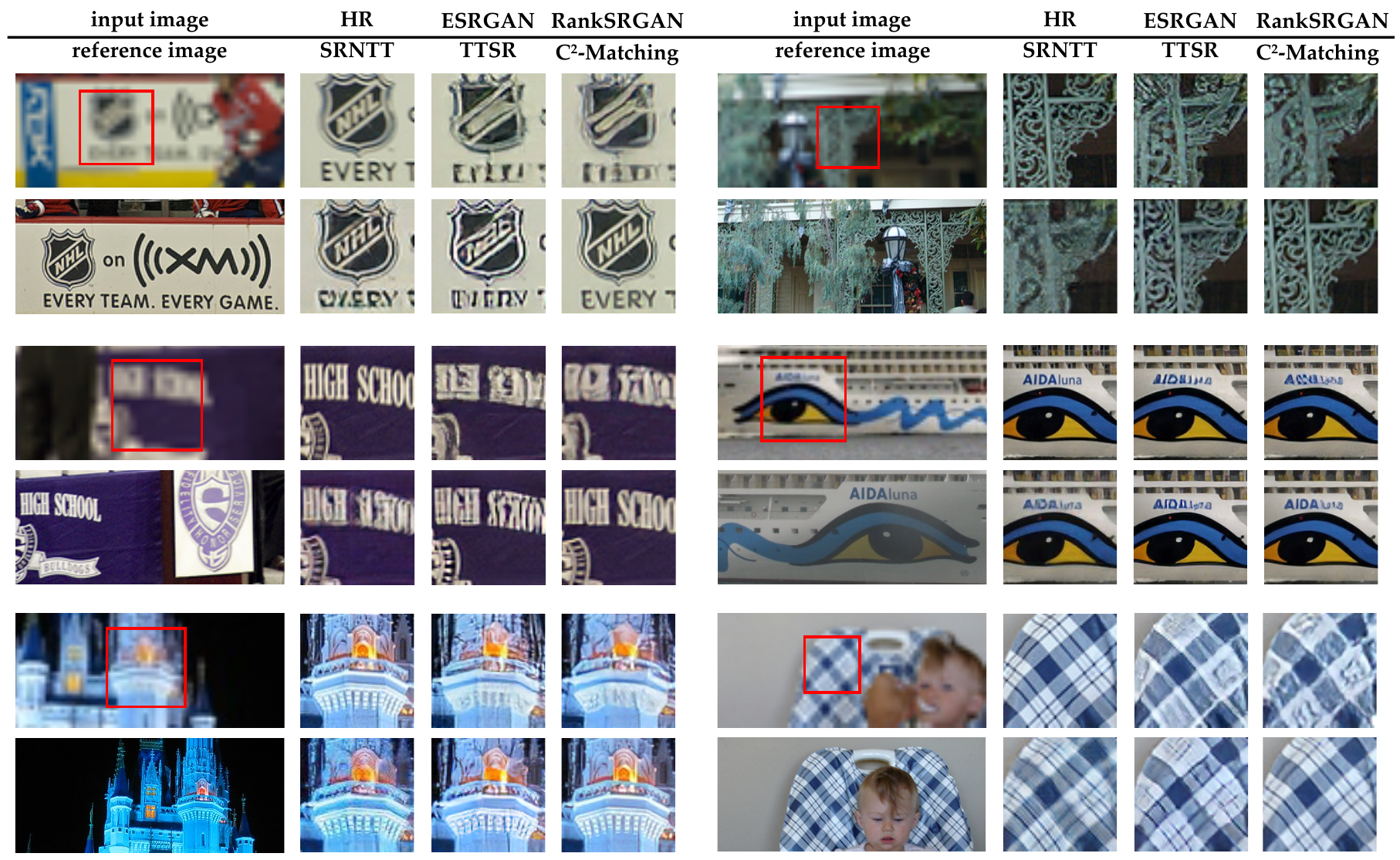}
  \end{center}
  \vspace{-8pt}
  \caption{\textbf{Qualitative Comparisons.} We compare our results with ESRGAN \cite{wang2018esrgan}, RankSRGAN \cite{zhang2019ranksrgan}, SRNTT \cite{zhang2019image}, and TTSR \cite{yang2020learning}. All these methods are trained with GAN loss. Our results have better visual quality with more texture details.}
  \label{visual_comp}
\end{figure*}

\section{Experiments}

\subsection{Datasets and Evaluation Metrics for Image SR}

\noindent\textbf{Training Dataset.} We train our models on the training set of CUFED5 dataset \cite{zhang2019image}, which contains 11,871 image pairs and each image pair has an input image and a reference image.

\noindent\textbf{Testing Datasets.} The performance are evaluated on the testing set of CUFED5 dataset, SUN80 dataset \cite{sun2012super}, Urban100 dataset \cite{huang2015single}, Manga109 dataset \cite{matsui2017sketch} and our WR-SR dataset. The CUFED5 dataset has 126 input images and each has 5 reference images with different similarity levels. The SUN80 dataset has 80 images with 20 reference images for each input image. WR-SR dataset has been introduced in Section~\ref{section_dataset}. As for the SISR datasets, we adopt the same evaluation setting as previous studies \cite{zhang2019image, yang2020learning}. The Urban100 dataset contains 100 building images, and the LR versions of images serve as reference images. The Manga109 dataset has 109 manga images and the reference images are randomly selected from the dataset.

\begin{figure*}[t]
  \begin{center}
      \includegraphics[width=1.0\linewidth]{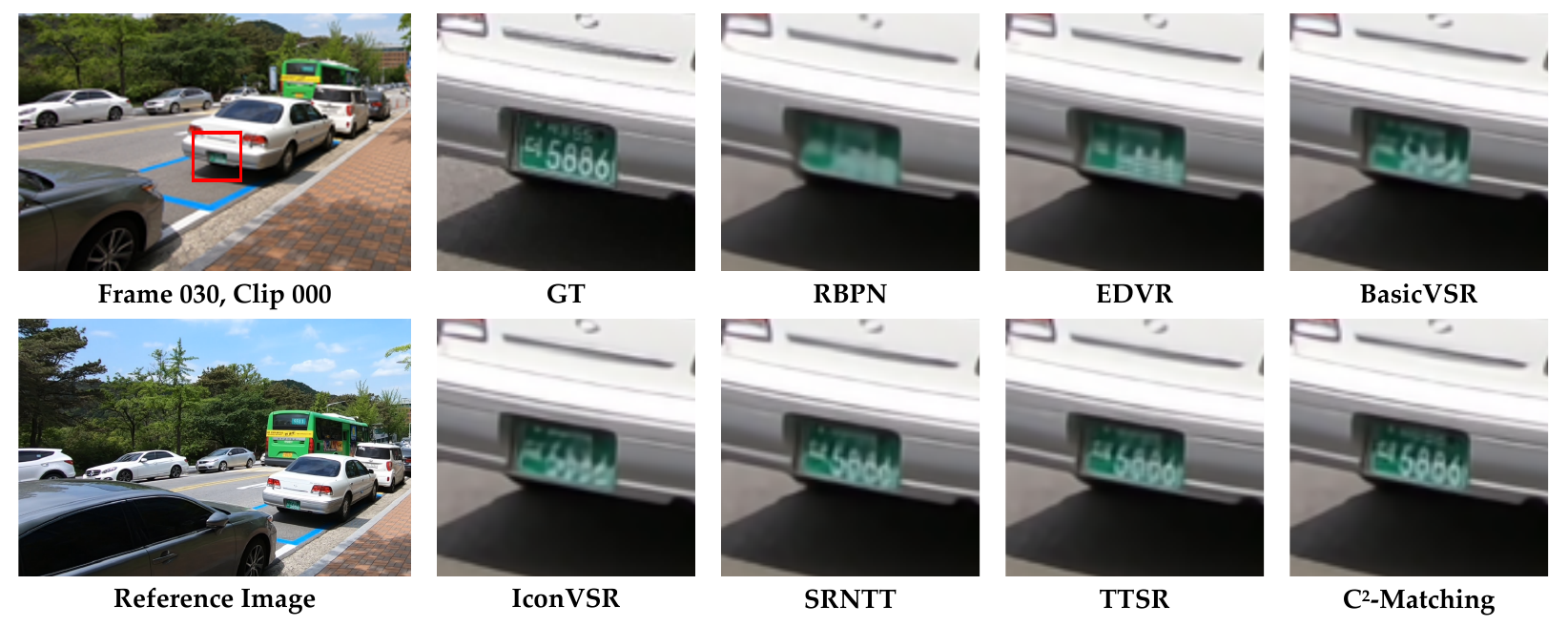}
  \end{center}
  \vspace{-8pt}
  \caption{\textbf{Qualitative Comparisons of Ref VSR task.} Our $C^{2}$-Matching can recover more details and sharper edges.}
  \label{visual_comp_video}
\end{figure*}

\begin{table*} 
\centering
	\caption{\small{Quantitative Comparisons on Ref VSR task (PSNR / SSIM). The results are computed on RGB channel for REDS4 datset and Y-Channel for Vid4 dataset.}}
\vspace{-6pt}
\addtolength{\tabcolsep}{6pt}
\small{
\begin{tabular}{l|cccc|c|c}
\Xhline{1pt}
\multirow{2}{*}{Method} & \multicolumn{5}{c|}{REDS4} & \multirow{2}{*}{Vid4} \\ \cline{2-6}
 & Clip 000 & Clip 011 & Clip 015 & Clip 020 & Average\\ \hline \hline
Bicubic & 24.55 / .649 & 26.06 / .726 & 28.52 / .803 & 25.41 / .739 &  26.14 / .729 & 23.79 / .635 \\ 
TOFlow \cite{xue2019video} & 26.52 / .754 & 27.80 / .786 & 30.67 / .861 & 26.92 / .795 & 27.98 / .799 & 25.89 / .765
\\
DUF \cite{jo2018deep} & 27.30 / .794 & 28.38 / .806 & 31.55 / .885 & 27.30 / .816 & 28.63 / .825 & - \\
RBPN \cite{haris2019recurrent} & 27.54 / .807 & 30.67 / .858 & 33.26 / .910 & 28.88 / .861 & 30.09 / .859 & 27.09 / .820 \\
EDVR-M \cite{wang2019edvr} & 27.75 / .815 & 31.29 / .873 & 33.48 / .913 & 29.59 / .878 & 30.53 / .870 & 27.10 / .819 \\
EDVR-L \cite{wang2019edvr} & 28.01 / .825 & 32.17 / .886 & 34.06 / .921 & 30.09 / .888 & 31.09 / .880 & 27.35 / .826 \\
PFNL \cite{yi2019progressive} & 27.43 / .801 & 29.74 / .842 & 32.55 / .899 & 28.52 / .851 & 29.63 / .850 & 26.69 / .802 \\
MuCAN \cite{li2020mucan} & 27.99 / .822 & 31.84 / .880 & 33.90 / .917 & 29.78 / .881 & 30.88 / .875 & - \\
BasicVSR \cite{chan2021basicvsr} & 28.40 / .843 & 32.47 / .898 & 34.18 / .922 & 30.63 / .900 & 31.42 / .891 & 27.24 / .825 \\
IconVSR \cite{chan2021basicvsr} & 28.54 / .848 & 32.88 / .902 & 34.51 / .927 & 30.76 / .902 & 31.67 / .895 & 27.40 / .828 \\ \hline \hline
SRNTT \cite{zhang2019image} & 28.76 / .854 & 32.91 / .904 & 34.69 / .929 & 30.90 / .905  & 31.82 / .898 & 28.43 / .878 \\
TTSR \cite{yang2020learning} & 28.81 / .854 & 32.98 / .903 & 34.76 / .930 & 30.92 / .905  & 31.85 / .898 &  28.75 / .887 \\
MASA-SR \cite{lu2021masa} & 28.76 / .854 & 32.93 / .904 & 34.75 / .931 & 30.89 / .905  & 31.83 / .898 & 28.60 / .886 \\
$C^{2}$-Matching & \textbf{28.98} / \textbf{.860} & \textbf{33.21} / \textbf{.908} & \textbf{34.91} / \textbf{.931} & \textbf{31.09} / \textbf{.907} & \textbf{32.05} / \textbf{.901} & \textbf{28.87} / \textbf{.896} \\
\Xhline{1pt}
\end{tabular}}
\label{quan_comp_video}
\end{table*}

\noindent\textbf{Evaluation Metrics.} PSNR and SSIM on Y channel of YCrCb space are adopted as evaluation metrics.
Input LR images for evaluation are bicubic downsampled $4\times$ from HR images.

\subsection{Results Comparisons on Image SR}
\noindent\textbf{Quantitative Comparison.} We compare the proposed $C^{2}$-Matching with representative SISR methods and Ref-SR methods. For SISR methods, we include SRCNN \cite{dong2015image}, EDSR \cite{lim2017enhanced}, RCAN \cite{zhang2018image}, SRGAN \cite{ledig2017photo}, ENet \cite{sajjadi2017enhancenet}, ESRGAN \cite{wang2018esrgan} and RankSRGAN \cite{zhang2019ranksrgan}. For Ref-SR methods, CrossNet \cite{zheng2018crossnet}, SRNTT \cite{zhang2019image}, SSEN \cite{Shim_2020_CVPR}, TTSR \cite{yang2020learning}, E2ENT$^{2}$ \cite{xiefeature}, CIMR \cite{yantowards} and MASA-SR \cite{lu2021masa} are included.

Table~\ref{quan_comp} shows the quantitative comparison results. We mark the methods trained with GAN loss in gray. 
On the standard CUFED5 benchmark, our proposed method outperforms state of the arts by a large margin. We also achieve state-of-the-art results on the Sun80 \cite{sun2012super}, Manga109 \cite{matsui2017sketch} and WR-SR dataset, suggesting the generalizability of $C^{2}$-Matching. On Urban100 dataset \cite{huang2015single}, our method obtains comparable results with MASA \cite{lu2021masa}.
Notably, CIMR \cite{yantowards} is a multiple reference-based SR method that transfers HR textures from a collection of reference images.
The performance of $C^{2}$-Matching given one reference image is better than that of CIMR when multiple reference images are provided, demonstrating the superiority of our method.

\noindent\textbf{Qualitative Evaluation.} Fig.~\ref{visual_comp} shows the qualitative comparison with state of the arts. We compare our method with ESRGAN \cite{wang2018esrgan}, RankSRGAN \cite{zhang2019ranksrgan}, SRNTT \cite{zhang2019image}, and TTSR \cite{yang2020learning}. The results of our method have the best visual quality containing many realistic details and are closer to their respective HR ground-truths. As shown in the top left example, $C^{2}$-Matching successfully recovers the exact word ``EVERY'' while other methods fail.

\begin{figure*}
  \begin{center}
      \includegraphics[width=1.0\linewidth]{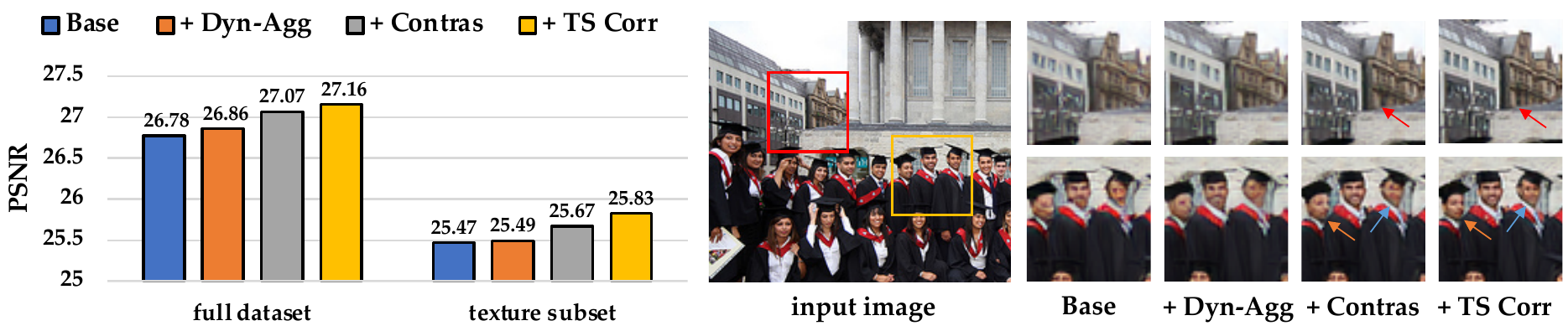}
  \end{center}
  \vspace{-10pt}
  \caption{\textbf{Ablation Study.} We evaluate the effectiveness of each component on the full CUFED5 dataset and the texture region subset. Since the contrastive correspondence (Contras) network and teacher-student correlation distillation (TS Corr) focus on improving texture details, we also add visual comparisons with the component added.}
  \label{ablation_study}
\end{figure*}

\subsection{Dataset and Evaluation Metrics for Video SR}

\noindent\textbf{Dataset.} REDS dataset \cite{nah2019ntire} contains 270 video clips. 
Each clip contains 100 frames subsampled from the high frame-rate counterpart containing 500 frames each.  
Following Wang \etal \cite{wang2019edvr}, REDS4 dataset are adopted as test set and the rest of clips are used for training. The first frame is selected as the HR reference image for each video clip (The first frame does not overlap with the test clip as it is from the high frame-rate clip.). We also evaluate the performance on Vid4 dataset \cite{liu2013bayesian}. For Vid4 dataset, the first frame is selected as the reference image. We will discuss the influence of the selection of reference images in Section~\ref{sec_futher_analysis}.

\noindent\textbf{Evaluation Metrics.} Input LR videos for evaluation are bicubic downsampled by $4\times$ from HR frames.
We evaluate PSNR and SSIM on RGB channels for REDS4 dataset \cite{nah2019ntire} and Y-channel for Vid4 dataset \cite{liu2013bayesian}. When computing evaluation metrics for Vid4 dataset, the first frame of each clip is excluded as it is selected as the reference image.

\subsection{Results Comparisons on Video SR}
\noindent\textbf{Quantitative Evaluation.}
We compare our proposed $C^{2}$-Matching with TOFlow \cite{xue2019video}, DUF \cite{jo2018deep}, RBPN \cite{haris2019recurrent}, EDVR \cite{wang2019edvr}, PFNL \cite{yi2019progressive}, MuCAN \cite{li2020mucan}, BasicVSR \cite{chan2021basicvsr} and IconVSR \cite{chan2021basicvsr}. We also adapt Reference-based Image SR methods SRNTT \cite{zhang2019image}, TTSR \cite{yang2020learning}, and MASA-SR\cite{lu2021masa} to the Ref VSR task. The quantitative comparison is shown in Table~\ref{quan_comp_video}. Our proposed $C^{2}$-Matching outperforms state-of-the-art VSR methods and also is superior to Ref VSR methods.

\noindent\textbf{Qualitative Evaluation.}
Fig.~\ref{visual_comp_video} shows the visual comparisons with state-of-the-art VSR methods and other Ref-VSR methods. With the aid of the reference image, Ref-VSR methods (SRNTT \cite{zhang2019image}, TTSR \cite{yang2020learning} and our proposed $C^{2}$-Matching) successfully recover the details while VSR methods (RBPN \cite{haris2019recurrent}, EDVR \cite{wang2019edvr}, BasicVSR and IconVSR \cite{chan2021basicvsr}) fail. Compared to SRNTT \cite{zhang2019image} and TTSR \cite{yang2020learning}, our proposed $C^{2}$-Matching can generate results with sharper edges.

\subsection{Ablation Study on Image SR}

We perform ablation studies to assess the effectiveness of each module. 
To evaluate the effectiveness of our proposed modules on texture regions, we select a subset of CUFED5 dataset that contains images with complicated textures; we name it ``texture subset".
On top of the base model without any proposed modules, we progressively add the dynamic aggregation module, contrastive correspondence network and teacher-student correlation distillation to show their effectivenesses. The results are shown in Fig.~\ref{ablation_study}.

\noindent\textbf{Dynamic Aggregation Module.} We first analyze the effectiveness of the Dynamic Aggregation (Dyn-Agg) module because it deals with the reference texture transfer problem. Only with a better texture transfer module would the improvements of correspondence matching module be reflected, \ie the base model accompanied with Dyn-Agg module provides a stronger model. The Dyn-Agg module dynamically fuses the HR information from reference images. Compared to the previous scheme \cite{zhang2019image, yang2020learning} that cropped patches of a fixed size from HR reference features, the Dyn-Agg module can flexibly aggregates HR reference features with learnable offsets $\Delta p_k$. With Dyn-Agg module, we observe an increment in PSNR by 0.08dB in the full dataset. 

\noindent\textbf{Contrastive Correspondence Network.} We further replace the fixed VGG feature matching module with the contrastive correspondence (Contras) network. With the learnable contrastive correspondence network, the PSNR value increases by about 0.2dB.
This result demonstrates the contrastive correspondence network computes more accurate correspondences and further boosts the performance of restoration. Fig.~\ref{ablation_study} shows one example of visual comparisons. With the contrastive correspondence network, the output SR images show more realistic textures.

\noindent\textbf{Teacher-Student Correlation Distillation.} With the teacher-student correlation (TS Corr) distillation, the performance further increases by 0.09dB on the whole dataset. For the texture subset, the performance increases by 0.16dB. The TS Corr module aims to push closer the correspondence of LR-HR student Contras network and that of HR-HR teacher Contras network. Since HR-HR teacher matching model is more capable of matching texture regions, the TS Corr module mainly boosts the performance on texture regions. As indicated in Fig.~\ref{ablation_study}, with TS Corr module, the textures of output SR images are enriched.

\begin{table*}[h] 
\centering
    \caption{\small{Ablation Study on Reference Video SR. `Optical Flow' refers to the model replacing the $C^{2}$-Matching with optical flows to align the Ref to LR. `w/o Attention' refers to the model with the attention module removed.}}
\addtolength{\tabcolsep}{6pt}
\small{
\begin{tabular}{l|cccc|c}
\Xhline{1pt}
\multirow{2}{*}{Method} & \multicolumn{5}{c}{REDS4} \\ \cline{2-6}
 & Clip 000 & Clip 011 & Clip 015 & Clip 020 & Average\\ \hline \hline
Optical Flow & 28.72 / .851 & 33.07 / .905 & 34.76 / .929 & 30.86 / .903 & 31.85 / .897  \\ 
w/o Attention & 28.91 / .858 & 33.14 / .906 & 34.78 / .929 & 31.02 / .906 & 31.96 / .900 \\ 
$C^{2}$-Matching & \textbf{28.98} / \textbf{.860} & \textbf{33.21} / \textbf{.908} & \textbf{34.91} / \textbf{.931} & \textbf{31.09} / \textbf{.907} & \textbf{32.05} / \textbf{.901} \\
\Xhline{1pt}
\end{tabular}}
\label{abalation_video}
\end{table*}

\begin{figure*}
  \begin{center}
      \includegraphics[width=1.0\linewidth]{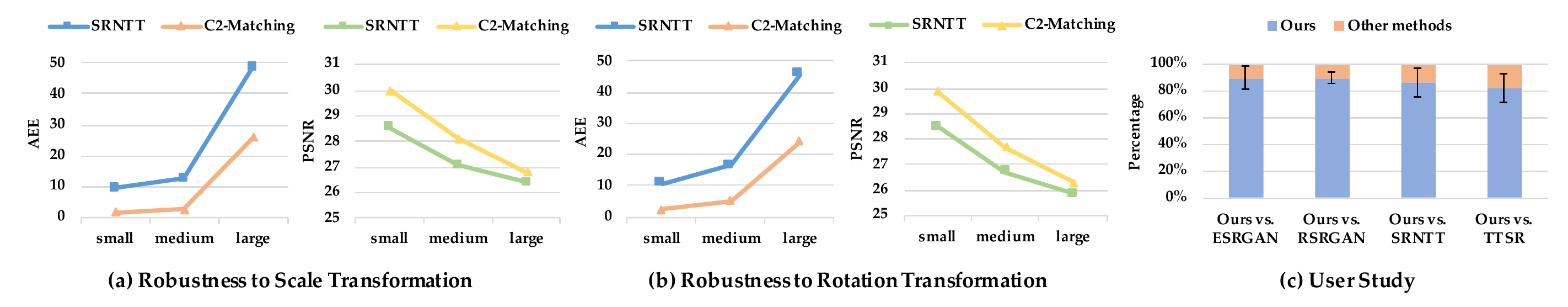}
  \end{center}
  \caption{\textbf{Further Analysis.} (a) Robustness to scale transformation. (b) Robustness to rotation transformations. The proposed $C^{2}$-Matching is more robust to scale and rotation transformation compared to SRNTT. (c) User study. Compared to other state of the arts, over $80\%$ users prefer our results.}
  \label{further_analysis}
\end{figure*}

\subsection{Ablation Study on Video SR}
\noindent\textbf{Necessity of $C^{2}$-Matching Module for Correspondence Matching in Ref VSR.}
In the Ref VSR task, LR frames can be regarded as low-resolution observations of the frames near the reference image. A straightforward way to align the reference frame to LR is to use optical flows. We perform an ablation study by replacing $C^{2}$-Matching with optical flows to compute the correspondence. 
Following BasicVSR~\cite{chan2021basicvsr}, 
we use the pretrained SPyNet~\cite{ranjan2017optical} to estimate optical flows 
for its simplicity and efficiency.
The quantitative results are shown in Table~\ref{abalation_video}. Using optical flows for alignment results in a performance drop by 0.2dB. 
The performance drop is due to the following two reasons: 1) Optical flows are known to be inaccurate in cases of large motions, and therefore performance could deteriorate for LR frames far from the reference; 2) Optical flows computed between LR frames are inevitably inaccurate, limiting the performance when they are applied to align the HR reference image. This ablation study validates the effectiveness of our proposed $C^{2}$-Matching for correspondence matching.

\noindent\textbf{Attention Mechanism.}
We use the attention mechanism to fuse the features from the forward propagation branch, backward propagation branch, and $C^{2}$-Matching branch. As shown in Table~\ref{abalation_video}, 
the performance drops by around 0.1dB without the attention mechanism. The experimental results demonstrate that the attention fusion brings additional improvements on top of the $C^{2}$-Matching module.

\subsection{Further Analysis}
\label{sec_futher_analysis}
\noindent\textbf{Robustness to Scale and Rotation Transformations.} We perform further analysis on the robustness of our $C^{2}$-Matching to scale and rotation transformations. We build a transformation-controlled dataset based on CUFED5 dataset. The scaled and rotated versions of input images serve as reference images. 
We adopt two metrics to measure the robustness: Average End-to-point Error (AEE) for matching accuracy and PSNR for restoration performance.

Figure~\ref{further_analysis} shows the robustness to the scale and transformations in AEE and PSNR. We separately analyze the impact of scale and rotation. We classify the degrees of scale and rotations into three groups: small, medium and large. The AEE rises as the degrees of transformations increases, suggesting that a larger degree of transformation makes correspondence matching harder. As shown by the AEE results in Fig.~\ref{further_analysis}(a, 
b), $C^{2}$-Matching computes more accurate correspondences than SRNTT under scale and rotation transformations. With the features that are more robust to scale and rotation transformations, according to the PSNR, the restoration performance of our proposed $C^{2}$-Matching is also more robust than that of SRNTT \cite{zhang2019image}. It should be noted that large transformations are not included during training but our proposed $C^{2}$-Matching still exhibits superior performance compared to SRNTT.

\noindent\textbf{User Study.} We perform a user study to further demonstrate the superiority of our method qualitatively. A total of 20 users were asked to compare the visual quality of our method and state of the arts on the CUFED5 dataset, including ESRGAN \cite{wang2018esrgan}, RankSRGAN \cite{zhang2019ranksrgan}, SRNTT \cite{zhang2019image} and TTSR \cite{yang2020learning}. We presented images in pairs, of which one was the result of our method, and asked users to choose the one offering better visual quality. As shown in Fig.~\ref{further_analysis}(c), over 80\% of the users felt that the result of our method is superior compared to that of baselines.

\noindent\textbf{Performance vs. Model Size Trade-Off.} The comparison of model size (\ie the number of trainable parameters) is shown in Table~\ref{table_param_comp}. $C^{2}$-Matching has a total number of 8.9M parameters and achieves a PSNR of 28.24dB. For a fair comparison in terms of model size, we build a light version of $C^{2}$-Matching, which has fewer trainable parameters. The $C^{2}$-Matching-$light$ is built by setting the number of the third and the fourth residual blocks to 8 and 4, respectively, and removing the first dynamic aggregation module. The $C^{2}$-Matching-$light$ has a total number of 4.8M parameters. The light version has fewer parameters than TTSR \cite{yang2020learning} but significantly better performance.

\noindent\textbf{Influence of Similarities of Reference Images on Performance of Ref VSR.} We study the restoration performance (in terms of PSNR) under different reference images with different similarities. For each clip in REDS dataset \cite{nah2019ntire}, the last 50 frames (51 - 100 frames) are used as the new video clip for evaluation. The 50th frame and the 21st frame are adopted as the very similar reference image and similar reference image, respectively. For the irrelevant reference image, we choose one frame from
another irrelevant clip. 
We also test the performance without any reference images using BasicVSR~\cite{chan2021basicvsr}.
The results are shown in Table~\ref{table_similarity}.
As the similarity of reference images increases, the performance of our proposed $C^{2}$-Matching becomes better. If an irrelevant reference image is given, the proposed $C^{2}$-Matching degenerates to a vanilla video SR method, such as BasicVSR in this work.

\noindent\textbf{Comparisons on Inference Time of $C^{2}$-Matching and Optical Flows in Ref VSR.}
In Table~\ref{abalation_video}, we validate that using $C^{2}$-Matching to align the reference frame to LR outperforms the variant of using optical flow by 0.2dB. Here, we compare the inference time of the $C^{2}$-Matching and Optical Flows. We evaluate the inference time on 100-frame videos with HR resolution of $1280 \times 720$.
The upsampling factor is $4$.
We test the inference time using one NVIDIA Tesla V100.
The inference time is obtained by averaging the total inference time of 100 videos.
The inference time of $C^{2}$-Matching is 163.11s per video, while the inference time of optical flow is 82.44s per video. 
The inference time is longer as it involves pixel-wise similarity matching between two feature maps, which takes up most of the time.
In future works, some mechanisms for accelerating similarity matching, such as PatchMatch~\cite{barnes2009patchmatch} and MASA-SR~\cite{lu2021masa}, can be applied to speed up the inference of $C^{2}$-Matching.

\begin{table}
\centering
\caption{\small{Model sizes of different methods.} Our proposed $C^{2}$-Matching-$light$ has a comparable number of parameters with SRNTT \cite{zhang2019image} but significant better performance. }
\addtolength{\tabcolsep}{8pt}
\small{
\begin{tabular}{l|c|c}
\Xhline{1pt}
\textbf{Method}  & \textbf{Params} & \textbf{PSNR / SSIM} \\ \hline \hline
 RCAN \cite{zhang2018image} & 16M & 26.06 / .769 \\ \hline
 RankSRGAN \cite{zhang2019ranksrgan} & 1.5M     & 22.31 / .635 \\ \hline
 CrossNet \cite{zheng2018crossnet} & 33.6M & 25.48 / .764 \\ \hline
 SRNTT \cite{zhang2019image} & 4.2M & 26.24 / .784 \\ \hline
 TTSR \cite{yang2020learning} & 6.4M & 27.09 / .804 \\ \hline
 $C^{2}$-Matching-$light$ & 4.8M  & 28.14 / .839 \\ \hline
 $C^{2}$-Matching & 8.9M & 28.24 / .841 \\ \hline 
\Xhline{1pt}
\end{tabular}
}
\label{table_param_comp}
\end{table}

\begin{table}
\centering
\caption{\small{Influence of similarities of reference images on performance (PSNR) of Ref VSR.} As the similarity increases, our proposed $C^{2}$-Matching achieves better performance. } 
\small{
\begin{tabular}{c|c|c|c|c}
\Xhline{1pt}
 \textbf{Clip} & \textbf{No Ref} & \textbf{Irrelevant} & \textbf{Similar} & \textbf{Very Similar} \\ \hline \hline
 Clip 000 & 28.16 & 28.15 & 28.30 & 28.88  \\ \hline
 Clip 011 & 32.61 & 32.52 & 32.73 & 33.33  \\ \hline
 Clip 015 & 34.41 & 34.44 & 34.54 & 35.05  \\ \hline
 Clip 020 & 30.41 & 30.39 & 30.53 & 30.95  \\ \hline
 Average & 31.40 & 31.38 & 31.53 & 32.05 \\
\Xhline{1pt}
\end{tabular}
}
\label{table_similarity}
\end{table}

\section{Conclusion}

In this paper, we develop a novel $C^{2}$-Matching for robust reference-based super-resolution. We focus on improving the correspondence matching accuracy between LR input images and HR reference images. Through our experiments and analysis, we demonstrate that more accurate correspondences benefit the performance of the Ref Image SR task. We also show the possibility of extending $C^{2}$-Matching into Ref VSR tasks. 
We believe that $C^{2}$-Matching can serve as a strong baseline and component for other tasks, such as capturing long-term correspondences in VSR. The performance of the Ref-SR task can also be further improved in circumstances where extremely large differences between input and reference images exist. 

\noindent\textbf{Acknowledgement}.
This study is supported by NTU NAP, MOE AcRF Tier 1 (2021-T1-001-088), and under the RIE2020 Industry Alignment Fund – Industry Collaboration Projects (IAF-ICP) Funding Initiative, as well as cash and in-kind contribution from the industry partner(s).

{ \small
  \bibliographystyle{IEEEtran}
  \bibliography{section/egbib}
}

%
%
%
%
\vspace{-30pt}
\begin{IEEEbiography}[{\includegraphics[width=1in,height=1.25in,clip,keepaspectratio]{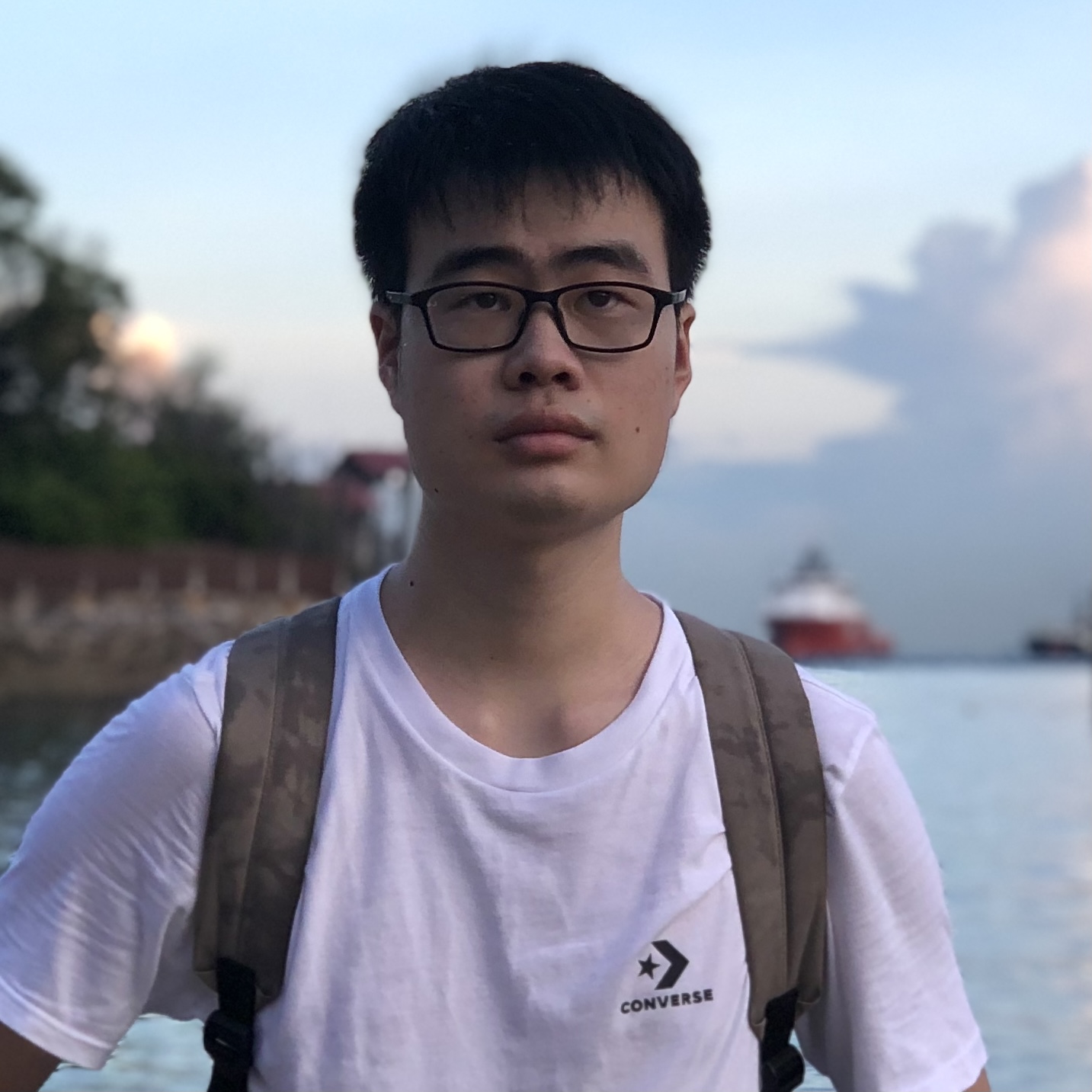}}]{Yuming Jiang}
  is currently a Ph.D. student at MMLab@NTU, Nanyang Technological University, supervised by Prof. Ziwei Liu and Prof. Chen Change Loy. He got his bachelor degree in computer science from Yingcai Honors College, University of Electronic Science and Technology of China (UESTC) in 2019. He received the Google PhD Fellowship in 2022. His research interests include image generation, manipulation and restoration.
\end{IEEEbiography}
\begin{IEEEbiography}[{\includegraphics[width=1in,height=1.25in,clip,keepaspectratio]{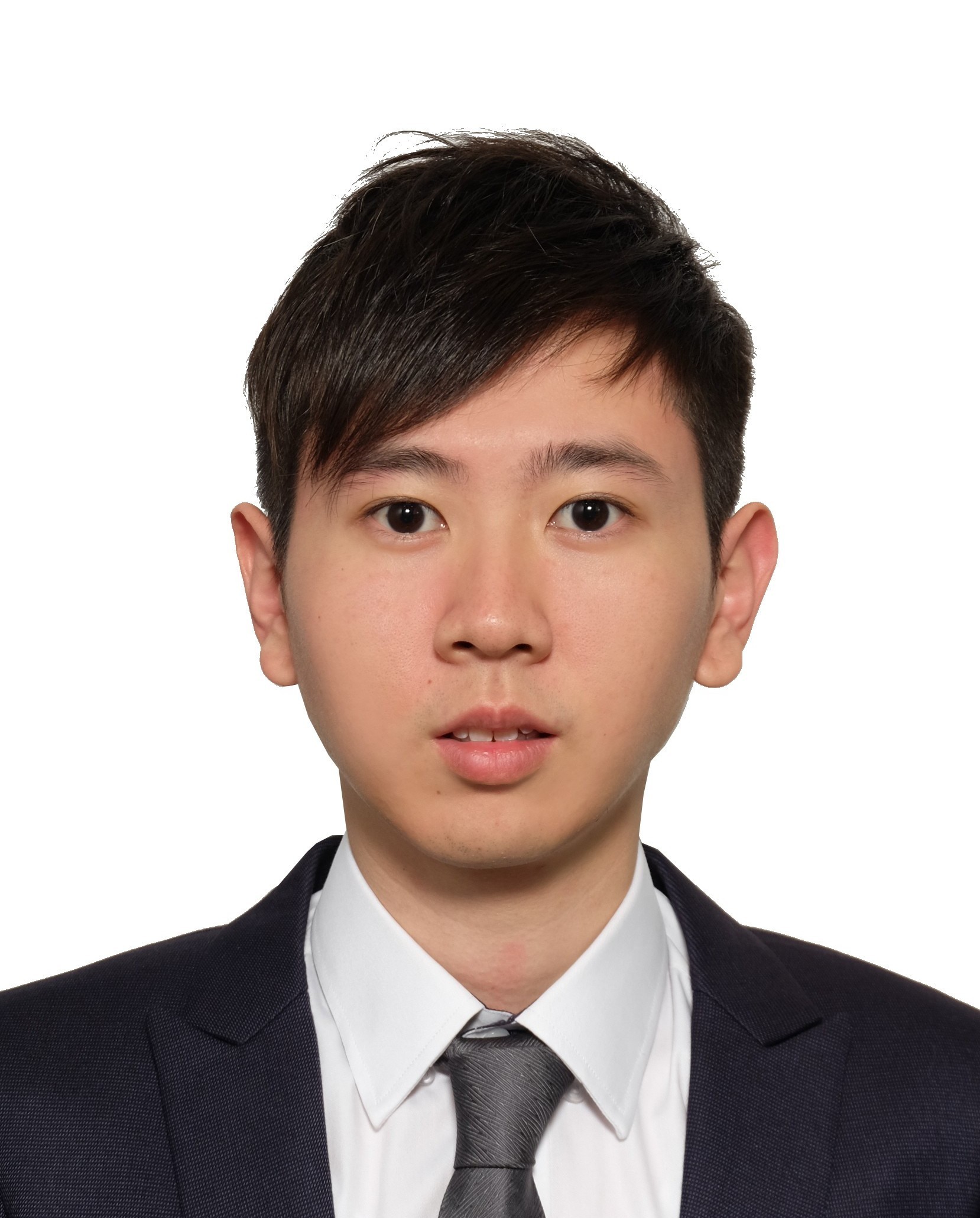}}]{Kelvin C.K. Chan}
  is currently a fourth-year PhD student at S-Lab, Nanyang Technological University. He received his MPhil degree in Mathematics as well as his BSc and BEng degrees from The Chinese University of Hong Kong. He was awarded the Google PhD Fellowship in 2021. He won the first place in multiple international challenges including NTIRE2019 and NTIRE2021, and was selected as an outstanding reviewer in ICCV 2021. His research interests include low-level vision, especially image and video restoration.
\end{IEEEbiography}
\begin{IEEEbiography}[{\includegraphics[width=1in,height=1.25in,clip,keepaspectratio]{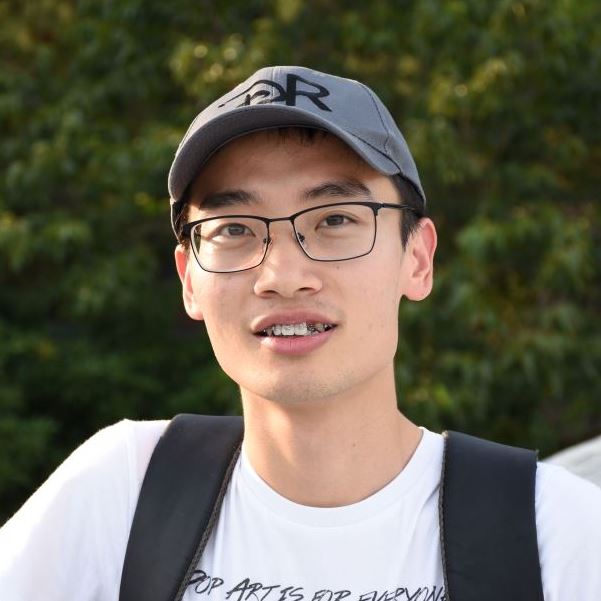}}]{Xintao Wang}
    is currently a senior researcher in Applied Research Center (ARC), Tencent PCG.
    He received his Ph.D. degree in the Department of Information Engineering, The Chinese University of Hong Kong, in 2020.
    He was selected as an outstanding reviewer in CVPR 2019 and an outstanding reviewer (honorable mention) in BMVC 2019.
    He won the first place in several international super-resolution challenges including NTIRE2019, NTIRE2018, and PIRM2018.
    His research interests include computer vision and deep learning, particularly focusing on image/video restoration and enhancement, generation and editing, etc.
\end{IEEEbiography}
\vspace{-30pt}

\begin{IEEEbiography}[{\includegraphics[width=1in,height=1.25in,clip,keepaspectratio]{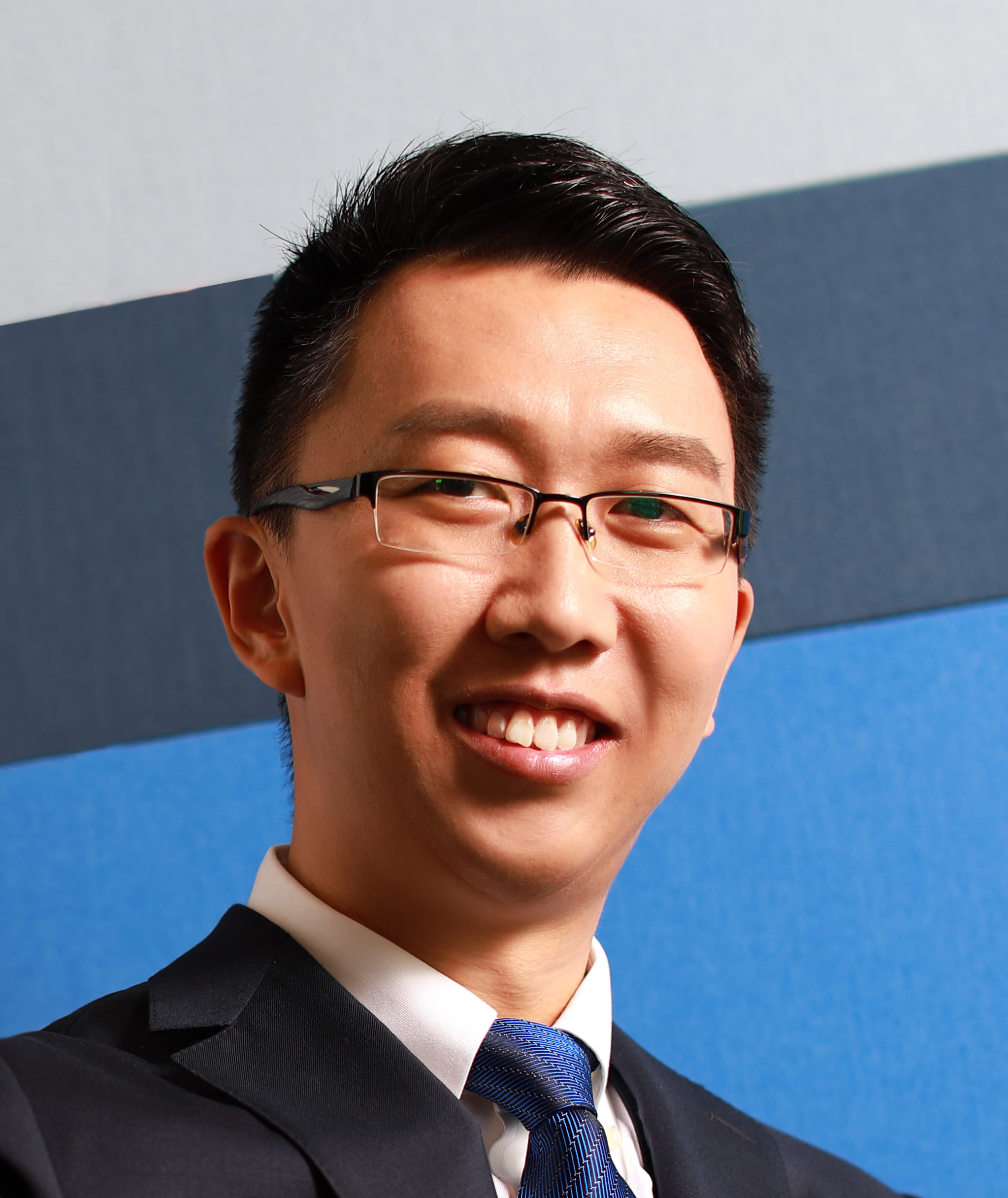}}]{Chen Change Loy} (Senior Member, IEEE) is an Associate Professor with the School of Computer Science and Engineering, Nanyang Technological University, Singapore. He is also an Adjunct Associate Professor at The Chinese University of Hong Kong. He received his Ph.D. (2010) in Computer Science from the Queen Mary University of London. Prior to joining NTU, he served as a Research Assistant Professor at the MMLab of The Chinese University of Hong Kong, from 2013 to 2018. He was a postdoctoral researcher at Queen Mary University of London and Vision Semantics Limited, from 2010 to 2013. He serves as an Associate Editor of the IEEE Transactions on Pattern Analysis and Machine Intelligence and International Journal of Computer Vision. He also serves/served as an Area Chair of major conferences such as ICCV 2021, CVPR 2021, CVPR 2019, and ECCV 2018. He is a senior member of IEEE. His research interests include image/video restoration and enhancement, generative tasks, and representation learning.
\end{IEEEbiography}
\vspace{-30pt}

\begin{IEEEbiography}[{\includegraphics[width=1in,height=1.25in,clip,keepaspectratio]{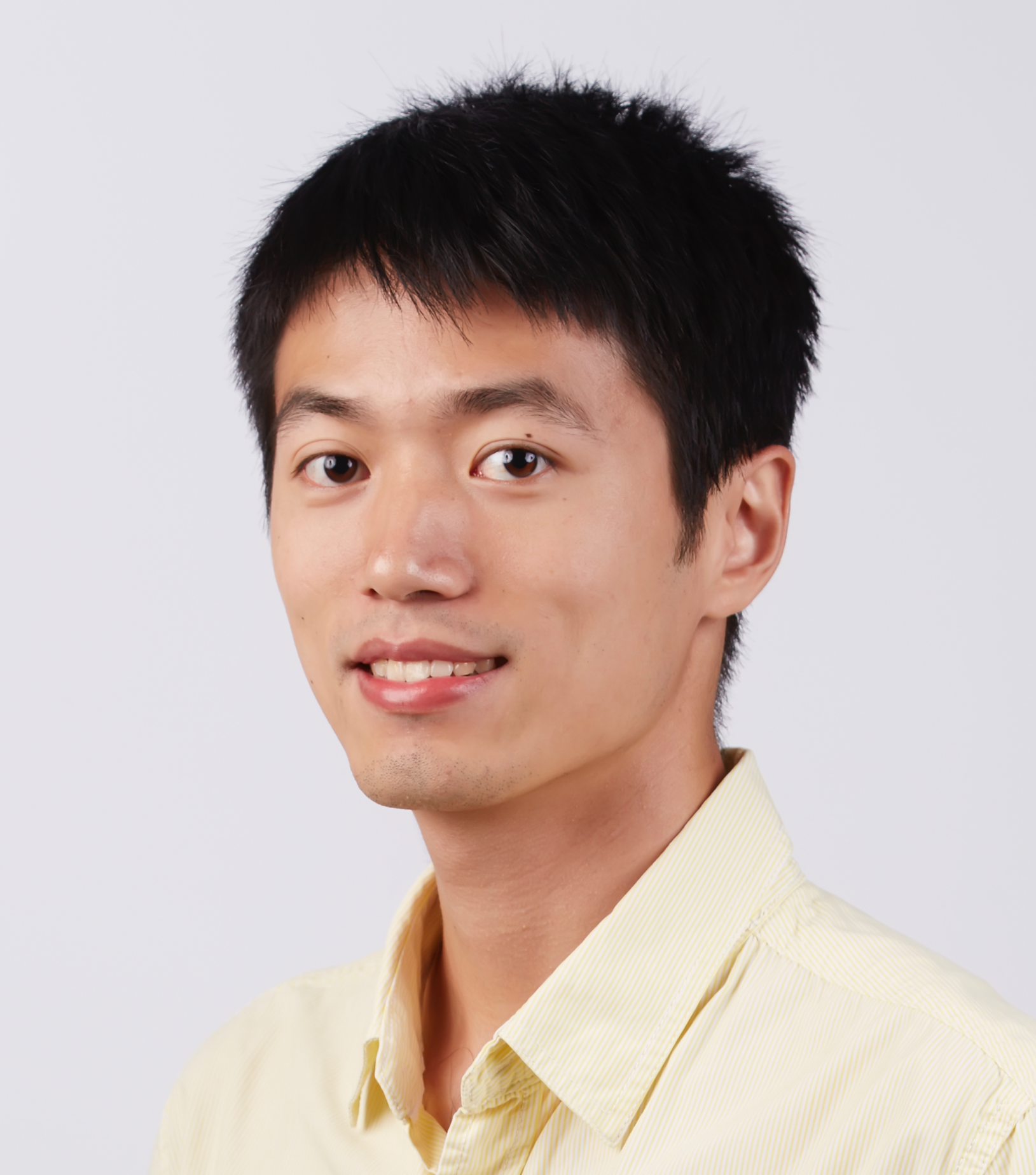}}]{Ziwei Liu} is currently a Nanyang Assistant Professor at Nanyang Technological University, Singapore. His research revolves around computer vision, machine learning and computer graphics. He has published extensively on top-tier conferences and journals in relevant fields, including CVPR, ICCV, ECCV, NeurIPS, ICLR, ICML, TPAMI, TOG and Nature - Machine Intelligence. He is the recipient of Microsoft Young Fellowship, Hong Kong PhD Fellowship, ICCV Young Researcher Award, HKSTP Best Paper Award and WAIC Yunfan Award. He serves as an Area Chair of CVPR, ICCV, NeurIPS and ICLR, as well as an Associate Editor of IJCV.
\end{IEEEbiography}




\end{document}